\documentclass[Afour,times,sageh]{sagej}

\usepackage{algpseudocode}
\usepackage{amsmath}
\usepackage{amssymb}
\usepackage{graphicx}
\usepackage{color}
\usepackage{epsfig} 
\usepackage{times} 
\usepackage{setspace}
\usepackage{mathtools}
\usepackage[normalem]{ulem}
\usepackage{array}
\usepackage{multirow}
\usepackage{booktabs}

\usepackage[ruled,linesnumbered]{algorithm2e}
\usepackage{subfig}

\usepackage[acronym]{glossaries}
\newacronym{ltl}{LTL}{Linear Temporal Logic}
\newacronym{gr1}{GR(1)}{generalized reactivity(1)}
\newacronym{pddl}{PDDL}{planning domain definition language}
\newacronym{ppddl}{PPDDL}{probabilistic planning domain definition language}

\newcommand{\precondset}[1][a]{\textrm{Pre}({#1})} 
\newcommand{\effectset}[2]{\textrm{Eff}^{#2}({#1})}  
\newcommand{\premask}[1][a]{\textrm{pre-mask}(#1)} 
\newcommand{\effmask}[2]{\textrm{eff-mask}^{#2}(#1)} 
\newcommand{\precondsym}[1][a]{\sigma_{\textrm{pre}(#1)}} 
\newcommand{\effectsym}[2]{\sigma^{\true}_{\textrm{eff}^{#2}(#1)}} 
\newcommand{\effectsymfalse}[2]{\sigma^{\false}_{\textrm{eff}^{#2}(#1)}} 
\newcommand{\sympremask}[1][a]{\Sigma_{\textrm{pre-mask}({#1})}} 
\newcommand{\effectsymstay}[2]{\sigma^{\textrm{stay}}_{\textrm{eff}^{#2}(#1)}} 
\newcommand{\anew}{a_\textrm{new}}

\newcommand{\envform}{\varphi_e}
\newcommand{\sysform}{\varphi_s}
\newcommand{\envinit}{\varphi_{\textrm{i}}^\textrm{e}}
\newcommand{\sysinit}{\varphi_{\textrm{i}}^\textrm{s}}
\newcommand{\envsafety}{\varphi_{\textrm{t}}^\textrm{e}}
\newcommand{\syssafety}{\varphi_{\textrm{t}}^\textrm{s}}
\newcommand{\envlive}{\varphi_{\textrm{g}}^\textrm{e}}
\newcommand{\syslive}{\varphi_{\textrm{g}}^\textrm{s}}
\newcommand{\fixed}{skills}
\newcommand{\tenv}{\tau_e}
\newcommand{\tsys}{\tau_s}
\newcommand{\tenvnew}{\tau_e^\textrm{new}}
\newcommand{\tsysnew}{\tau_s^\textrm{new}}
\newcommand{\tsyshard}{\tau_s^\textrm{hard}}
\newcommand{\syshard}{\varphi_{\textrm{t,hard}}^{\textrm{s}}}
\newcommand{\envsafetyfixed}{\varphi_{\textrm{t,\fixed}}^{\textrm{e}}}
\newcommand{\envlivetask}{\varphi_{\textrm{g,task}}^{\textrm{e}}}
\newcommand{\envinittask}{\varphi_{\textrm{i,task}}^{\textrm{e}}}
\newcommand{\syssafetyfixed}{\varphi_{\textrm{t,\fixed}}^{\textrm{s}}}
\newcommand{\syssafetytask}{\varphi_{\textrm{t,task}}^{\textrm{s}}}
\newcommand{\syslivetask}{\varphi_{\textrm{g,task}}^{\textrm{s}}}
\newcommand{\sysinittask}{\varphi_{\textrm{i,task}}^{\textrm{s}}}
\newcommand{\enveff}{\varphi_{\textrm{t,eff}}^{\textrm{e}}}
\newcommand{\envnoact}{\varphi_{\textrm{t,no\_act}}^{\textrm{e}}}
\newcommand{\envmxsyms}{\varphi_{\textrm{t,mx\_syms}}^{\textrm{e}}}

\newcommand{\syspre}{\varphi_{\textrm{t,pre}}^{\textrm{s}}}

\newcommand{\sysmxskills}{\varphi_{\textrm{t,mx\_skills}}^{\textrm{s}}}
\newcommand{\phifixed}{\varphi_{\textrm{\fixed}}}
\newcommand{\phitask}{\varphi_{\textrm{task}}}
\newcommand{\phifull}{\varphi_{\textrm{full}}}
\newcommand{\skills}{\mathcal{A}}

\newenvironment{flushitemize}{%
\begin{list}{$\bullet$}
   {\setlength{\leftmargin}{15pt}}%
    \setlength{\labelwidth}{20pt}
    \setlength{\itemindent}{0pt}
    \setlength{\labelsep}{0.5em}
 \setlength{\itemsep}{1pt}
 \setlength{\parskip}{0pt}
 \setlength{\parsep}{0pt}}
 {\end{list}}

\newcolumntype{s}{>{\centering\arraybackslash}p{30mm}}
\usepackage{makecell}

\definecolor{purple}{rgb}{0.5,0,0.5}

\newcommand{\set}[1]{\{#1\}}

\newcommand{\vars}{\mathcal{AP}}
\newcommand{\dom}{\mathcal{V}}
\newcommand{\obj}{\Phi}
\newcommand{\game}{G}

\newcommand{\winning}{Z}

\newcommand{\initstates}{\theta^{init}}

\newcommand{\inp}{\mathcal{E}}
\newcommand{\out}{\mathcal{S}}
\newcommand{\user}{\mathcal{R}}
\newcommand{\varinp}{v_{e}}
\newcommand{\varout}{v_{s}}
\newcommand{\varuser}{v_{u}}
\newcommand{\varstate}{v_{\vars}}
\newcommand{\learned}{\Sigma}
\newcommand{\counterstrat}{C_\textrm{c.s.}}
\newcommand{\strat}{C}
\newcommand{\blueperson}{BluePerson}
\newcommand{\greenperson}{GreenPerson}
\newcommand{\react}{React}
\newcommand{\switch}{Switch}
\newcommand{\factors}{F}

\newcommand{\setor}{~|~}

\newcommand{\true}{{\tt{True}}}
\newcommand{\false}{{\tt{False}}}

\glsdisablehyper
\begin{document}

\def\volumeyear{2021}

\title{Automatic Encoding and Repair of Reactive High-Level Tasks with Learned Abstract Representations}

\runninghead{Pacheck, James, Konidaris, Kress-Gazit}

\author{Adam Pacheck\affilnum{1}, Steven James\affilnum{2}, George Konidaris\affilnum{3}, Hadas Kress-Gazit\affilnum{1}}
\affiliation{\affilnum{1} Cornell University, Ithaca, NY, USA\\
\affilnum{2} University of the Witwatersrand, Johannesburg, South Africa\\
\affilnum{3} Brown University, Providence, RI, USA}
\corrauth{Adam Pacheck Cornell University, Ithaca, NY 14853, USA}
\email{akp84@cornell.edu}

\begin{abstract}

We present a framework that, given a set of skills a robot can perform, abstracts sensor data into symbols that we use to automatically encode the robot's capabilities in \gls{ltl}. 
We specify reactive high-level tasks based on these capabilities, for which a strategy is automatically synthesized and executed on the robot, if the task is feasible. 
If a task is not feasible given the robot's capabilities, we present two methods, one enumeration-based and one synthesis-based, for automatically suggesting additional skills for the robot or modifications to existing skills that would make the task feasible. 
We demonstrate our framework on a Baxter robot manipulating blocks on a table, a Baxter robot manipulating plates on a table, and a Kinova arm manipulating vials, with multiple sensor modalities, including raw images. 

\end{abstract}


\maketitle


\section{Introduction}
\glsresetall

Generally-useful robots will be required to generate intelligent behavior from high-level task specifications, especially if they are to be used by non-experts.
Robots should have the ability to reason about their actions (or \textit{skills}),  a task's goals and its constraints, and generate the behavior necessary to achieve the task, autonomously.
One promising formalism for describing tasks and skills is \gls{ltl}~\citep{pnueli1977temporal}.
\gls{ltl} allows one to encode (i) skills that have nondeterministic outcomes, (ii) safety constraints, (iii) reactive tasks, where the robot responds to the environment, and (iv) tasks with complex goals that go beyond reaching a goal state (as is typical in planning languages such as PDDL~\citep{mcdermott1998pddl}).
Furthermore, there exist different algorithms that enable a robot to synthesize a controller that is guaranteed to complete a specified task for fragments of \gls{ltl}, such as \gls{gr1}~\citep{bloem2012synthesis}.

However, writing \gls{ltl} specifications is not trivial. 
Since it is a discrete logic, it requires an abstraction of the problem.
Often, abstractions are handcrafted or constructed from a simplified model of the world which may not fully capture the outcomes of the robot's skills.
Recently, work has looked at creating abstractions directly from sensor data \citep{konidaris2018skills, jetchev13, mugan09, ugur15}, but those approaches have not been extended to seamlessly integrate with mission specification approaches that employ formal languages such as \gls{ltl}.

In approaches to synthesizing controllers from \gls{ltl} specifications, the robot skills and task are encoded as \gls{ltl} formulas and then the algorithms find a strategy such that the task is guaranteed to be achieved, if feasible \cite{kress2018synthesis}. 
However, if the task is not possible given the current skills of the robot, it is difficult to understand why, much less what needs to be done to make the task possible.
Recently, several methods for debugging \gls{ltl} specifications have been proposed (e.g. \citep{raman2013explaining,chatterjee2008environment,konighofer2009debugging}) along with methods for suggesting modifications to specifications (e.g. \citep{pacheck2019automatic, pacheck2020finding, fainekos2011revising, kim2015minimal}).
When debugging specifications, the user still needs to decide how to repair the specification.
In general, methods for finding suggestions that repair a specification often require changing the task, rather than changing the robot's skills to allow the robot to complete the task as specified.

Here, we build on the work in \citet{pacheck2019automatic} and \citet{pacheck2020finding}.
Extending \citet{pacheck2019automatic}, we demonstrate encoding robot skills using multiple types of sensor data, including raw images.
By creating abstractions and encoding skills directly from sensor data, we are able to take into account unmodeled nondeterminism in a robot's skills without having to hand design abstractions.
A user is then able to use these abstractions to specify a high-level task for the robot.
If the robot is always able to accomplish the task, we can use existing methods in \gls{ltl} to generate a strategy that will guarantee the robot will accomplish the task (e.g.  \citep{bloem2012synthesis}).
If the task is not possible, we propose two algorithms for repair---\textit{enumeration-based} and \textit{synthesis-based}---that will suggest additional skills or modifications to skills that would allow the robot to successfully accomplish the task as specified by the user.
We modify the enumeration-based repair algorithm of \citet{pacheck2019automatic} to allow for suggestions containing more than one skill and expand the synthesis-based repair algorithm of \citet{pacheck2020finding} to allow for the repair of reactive tasks, where the robot behavior depends on the behavior of the uncontrolled environment.
Furthermore, we demonstrate our approach on new tasks and a new physical robot (the Kinova arm in addition to the Baxter).

\textbf{Contributions}:
Given a set of skills a robot is able to perform, we present a framework that uses sensor data to automatically create an abstraction and encode skills in \gls{ltl} and then, given a task written as an \gls{ltl} formula over the abstraction, provides skill suggestions to repair infeasible tasks. 
Specifically, 
(a) We propose a method to automatically encode the robot capabilities into \gls{ltl}, directly from sensor data, which is then used to automatically synthesize high-level robot behaviors to accomplish a user-specified task. 
(b) If a user-specified task is not feasible due to a missing skill, we present two approaches---enumeration-based and synthesis-based---for automatically suggesting skills that repair the task (i.e. make it executable by the robot). 
(c) We demonstrate our approach on two physical systems: a Baxter robot manipulating blocks and pushing plates, and a Kinova arm manipulating vials.

\section{Related Work}
This work deals with abstractions, planning, synthesis, and repair.
Each of these areas is often dealt with individually, with overlap between some, but rarely all, of the areas.

\textbf{Abstraction Creation:} 
To enable robots to perform high-level tasks, the robot's capabilities, state, and environment are typically abstracted into predicates that include the robot's skills and their effects on the robot's state and environment. 
These predicates are often abstractions of the state space \citep{kress2018synthesis,Mazo:2010:PTE:2144310.2144372,Finucane2010}. 

There are several existing approaches to generating abstractions directly from low-level observations. 
These include learning symbols to model an agent's skills \citep{konidaris2018skills}, modelling an agent with parameterized actions \citep{ames2018learning}, and learning agent-centric symbols that can be transferred to new tasks in simple video games \citep{james2020}. 
\citet{ugur15,ugur15b} learn object-centric representations for a manipulation task. While their system can be used for symbolic planning on a physical robot, object features are specified prior to learning.
This approach is extended to learn representations directly from raw image data using a neural network \citep{ahmetoglu20}. In both cases, however, certain predicates are manually inserted to generate a sound representation.

\citet{jetchev13} learn relational symbols and operators directly from geometric data. 
However, the size of the search space is large, requiring one dimension for every parameter of every symbol, which restricts its ability to scale to large problems.
\citet{mugan09,mugan11} iteratively discretize a continuous state space to construct a model suitable for planning. Skills are then learned to reach these discretized states. 
This can be seen as a ``symbols-first'' approach, where skills are learned to achieve an initial discretization, which is then refined as necessary.

\citet{asai18} learn deterministic action operators directly from pixels using an autoencoder, where the bottleneck layer represents the set of propositions set to true and false. 
However, it is unclear how to extend the approach to the stochastic setting.
A similar approach produces deterministic object-centric representations \citep{asai19}, but the symbols are encoded implicitly and cannot be transformed into a language that can be used by existing planners.

While all of these approaches to generate abstraction can learn representations for planning, they lack a mechanism to correct a model that is imperfect or insufficient to solve a given task.
Recent works have begun to bridge the gap of modifying models that are insufficient to solve a task \citep{pacheck2019automatic, pacheck2020finding}.

\textbf{Planning: }
Planning algorithms and frameworks use abstractions to find a sequence of commands to reach a goal state \citep{fikes1971strips,fox2003pddl2,Ghallab2016}. 
If there is uncertainty in the outcome of skills, planners exist that will return the sequence of skills that is most likely to accomplish the task, but often require replanning when an unexpected effect occurs (e.g. \citep{Yoon2007}). 
When there is uncertainty in sensing, initial state, and actuation, conditional planning can return a plan that will take these into account \citep{Ghallab2016}.
If the robot cannot observe all of its environment, conformant planning generates a plan for a robot to accomplish its goal \citep{Ghallab2016}.
However, these goals are typically defined as a desired end state, while we consider more complex tasks. 
Additionally, if a planner fails to find a plan, to the best of our knowledge, planners are unable to suggest new skills that result in a valid plan.

\textbf{Synthesis: }Work in synthesis for robotics from temporal logic specifications~\citep{kress2018synthesis} allows us to specify a reactive high-level task for a robot and produce either a strategy guaranteed to succeed, or a proof that the task cannot be accomplished (e.g. \citep{lahijanian2012temporal,he2018automated,wongpiromsarn2010receding,decastro2016nonlinear,kress2009temporal,He2019}).

\begin{figure*}[t]
\centering
\includegraphics[width=\textwidth]{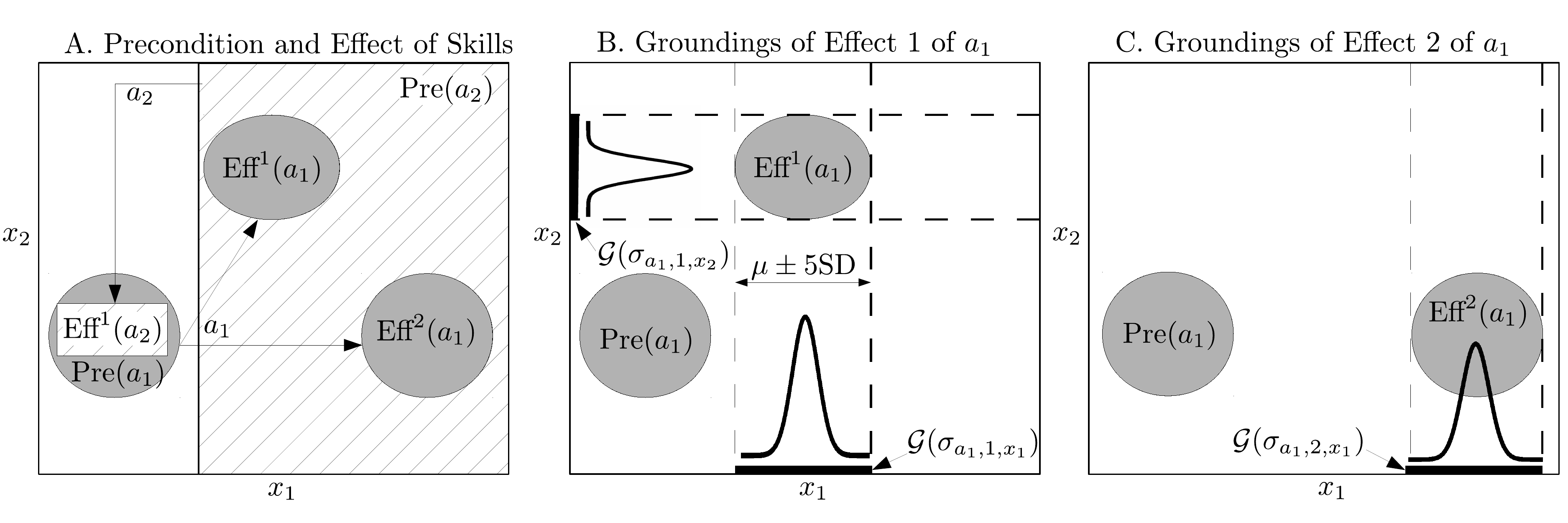}
\caption{Example demonstrating the symbol generation process. 
(A) Two skills $a_1$ and $a_2$ and their precondition and effect sets. 
(B,C) The grounding sets of the symbols generated from skill $a_1$. 
The robot needs to consider the value of both $x_1$ and $x_2$ when deciding if it can apply $a_1$, so $\premask[a_1] = \{\true,\true\}$. 
The robot only needs to consider the value of $x_1$ when deciding if it can apply $a_2$, so $\premask[a_2] = \{\true,\false\}$. 
The application of $a_1$ either changes $x_1$ and $x_2$ or only $x_1$, so $\effmask{a_1}{1} = \{\true,\true\}$ and $\effmask{a_1}{2} = \{\true,\false\}$. 
In effect 1 of $a_1$, $\effectsym{a_1}{1} = \{\sigma_{a_1,1,x_1},\sigma_{a_1,1,x_2}\}$ become $\true$ and $\effectsymfalse{a_1}{1} = \{\sigma_{a_1,2,x_1},\sigma_{a_2,1,x_1},\sigma_{a_2,1,x_2}\}$ become $\false$. 
In effect 2 of $a_1$, $\effectsym{a_1}{2} = \{\sigma_{a_1,2,x_1}\}$ becomes $\true$, $\effectsymfalse{a_1}{2} = \{\sigma_{a_1,1,x_1},\sigma_{a_2,1,x_1}\}$ becomes $\false$, and $\effectsymstay{a_1}{2} = \{\sigma_{a_1,1,x_2},\sigma_{a_2,1,x_2}\}$ do not change. Figure from \citet{pacheck2019automatic}.}
\label{fig:sym_gen}
\end{figure*}

\textbf{Specification Debugging: } 
Using synthesis, robots can find a strategy to accomplish a task that accounts for all possible outcomes of their skills and changes in the environment. 
However, if there does not exist a strategy to accomplish a task, it is difficult even for expert users to determine the cause, much less find a solution.
If completion of a task cannot be guaranteed, work has enabled synthesis algorithms to provide explanations as to what caused the problem (e.g. \citep{raman2013explaining,chatterjee2008environment,konighofer2009debugging}).
For fragments of \gls{ltl}, such as \gls{gr1}, synthesis algorithms can also produce counterstrategies that provide details on why the specification cannot be satisfied \citep{konighofer2009debugging}.
Tools such as Slugs \citep{ehlers2016slugs} exist that allow users to step through strategies and counterstrategies for debugging purposes.

\textbf{Specification Repair: }
Beyond providing methods to debug specifications, work has proposed methods to provide repair suggestions and automated fixes to be made to specifications \citep{alur2013counter,li2011mining}. 
Some frameworks allow for certain aspects of a specification to be skipped at runtime if they are not possible \citep{lahijanian2016iterative}.
In \citet{lahijanian2016iterative}, the task is split into safety constraints that must always be satisfied and liveness guarantees that should be satisfied if possible.
Work by \citet{fainekos2011revising} and \citet{kim2015minimal} also consider revising specifications.

These works (\citep{fainekos2011revising,kim2015minimal,alur2013counter,li2011mining,lahijanian2016iterative}) focus on restricting the behavior of the environment or modifying the goals of the robot to make the task possible.
In this work, we provide suggestions that extend the capabilities of the robot through additional skills or modifications to skills that are grounded in the sensor-based abstract representation. 
Instead of changing what we would like the robot to do, we give it additional capabilities that allow it to accomplish the desired task.

\section{Preliminaries}\label{sec:prelim}

\subsection{Skills}

We model the abilities of the robot as a set of skills, $\skills$, operating over a world with a continuous state space $(x_1,\ldots,x_n)\in X\subseteq \mathbb{R}^n$. 
Each skill $a \in \skills$ has a region from which it is applicable, termed the \textit{precondition} of $a$, $\precondset \subseteq X$. 
The application of $a$ will result in the state being in one of $j \in \{1,\ldots,k(a)\}$ possible \textit{effect sets}, denoted by $\effectset{a}{j} \subseteq X$. 
We introduce the example in Figure~\ref{fig:sym_gen} to illustrate the main ideas of skills and symbol generation. 
In this two-dimensional space, a robot has two skills $a_1$ and $a_2$ that allow it to move between regions, as shown by the arrows.
In Figure~\ref{fig:sym_gen}, $a_1$ has a nondeterministic outcome, resulting in either $\effectset{a_1}{1}$ or $\effectset{a_1}{2}$. 

\subsection{Symbol Generation}

The process of symbol generation \citep{konidaris2018skills} automatically constructs a set of symbols which are used for planning. 
The finite set of propositional symbols $\learned$ represents the effect sets of $a \in \skills$. 
Each $\sigma \in \learned$ is grounded via the \textit{grounding operator} $\mathcal{G}$ to the state space $X$.

The values of some $x_i$ may matter in determining whether a skill can be applied, while the values of others may not. 
We denote this in the \textit{precondition mask} of $a$, $\premask[a] \in \mathbb{B}^n$, where $\premask(i) = \true$ whether the value of $x_i$ influences if $a$ can be applied, and $\false$ otherwise. 
We create a classifier to test inclusion in $\precondset$, which is defined for $x_i$ for which $\premask[a](i) = \true$ . 
Similarly, when a skill is applied, it may change some or all of the state variables. 
We denote this in the \textit{effect mask}, $\effmask{a}{j} \in \mathbb{B}^n$, where $\effmask{a}{j}(i) = \true$ if the value of $x_i$ is modified by the application of $a$ in the $j^{th}$ outcome and $\false$ otherwise. 
In Figure \ref{fig:sym_gen}, to apply $a_1$, the values of both $x_1$ and $x_2$ matter, so $\premask[a_1] = \{\true,\true\}$. 
However, we need only consider the value of $x_1$ to determine if $a_2$ can be applied, so $\premask[a_2] = \{\true,\false\}$. 
Effect 1 of $a_1$ changes the value of $x_1$ and $x_2$ so $\effmask{a_1}{1} = \{\true,\true\}$, while effect 2 only changes the value of $x_1$, so $\effmask{a_1}{2} = \{\true,\false\}$.

We define factors $f_q \in F \subset 2^{X}$ that denote which state variables $x_i$ always change together.
A separate $\sigma$ is created\footnote{Note that $\sigma$ are only generated from effect sets in \citet{konidaris2018skills}.} for each $a$, $j$, and $f_q$ when $\effmask{a}{j}(i) = \true\ \forall x_i \in f_q$. 
We add subscripts to $\sigma$ and say each $\sigma_{a,j,f_q}$ grounds to a set over the state variables $x_i \in f_q$. 
We find the grounding by fitting either a Gaussian or using Kernel Density Estimation with a Gaussian kernel and consider $\mathcal{G}(\sigma_{a,j,x_i})$ to be the set of states spanned by five standard deviations from the mean.
If the raw data is found to be the same by a two-sample Kolmogorov-Smirnov test \citep{hollander1973nonparametric} or in the case of higher dimensions if the mean and variance are similar \citep{konidaris2018skills}, the two symbols are merged into one symbol.
The set of symbols referring to a single factor $f_q$ is $\Sigma_{f_q} = \{\sigma_{a,j,f_q}|a\in \skills, j\in \{1,\ldots,k(a)\}\}$. 
The set of all symbols is $\Sigma = \bigcup_{f_q\in F}\Sigma_{f_q}$. 
For readability, when $f_q = \set{x_i}$, we denote $\sigma_{a,j,f_q}$ as $\sigma_{a,j,x_i}$.
In Figure~\ref{fig:sym_gen}B, $\effectset{a_1}{1}$ results in two symbols, $\sigma_{a_1,1,x_1}$ and $\sigma_{a_1,1,x_2}$, because $\effmask{a_1}{1} = \{\true,\true\}$. 
Only one symbol, $\sigma_{a_1,2,x_1}$ is generated from $\effectset{a_1}{2}$ as $\effmask{a_1}{2} = \{\true,\false\}$.
In this example, all factors are singletons.

During the symbol generation process, skills that have different effects from different preconditions are partitioned into multiple skills \citep{konidaris2018skills}.

\subsection{Linear Temporal Logic (LTL)}\label{sec:ltl}

\begin{figure*}[t]
\centering
\includegraphics[width=0.9\textwidth]{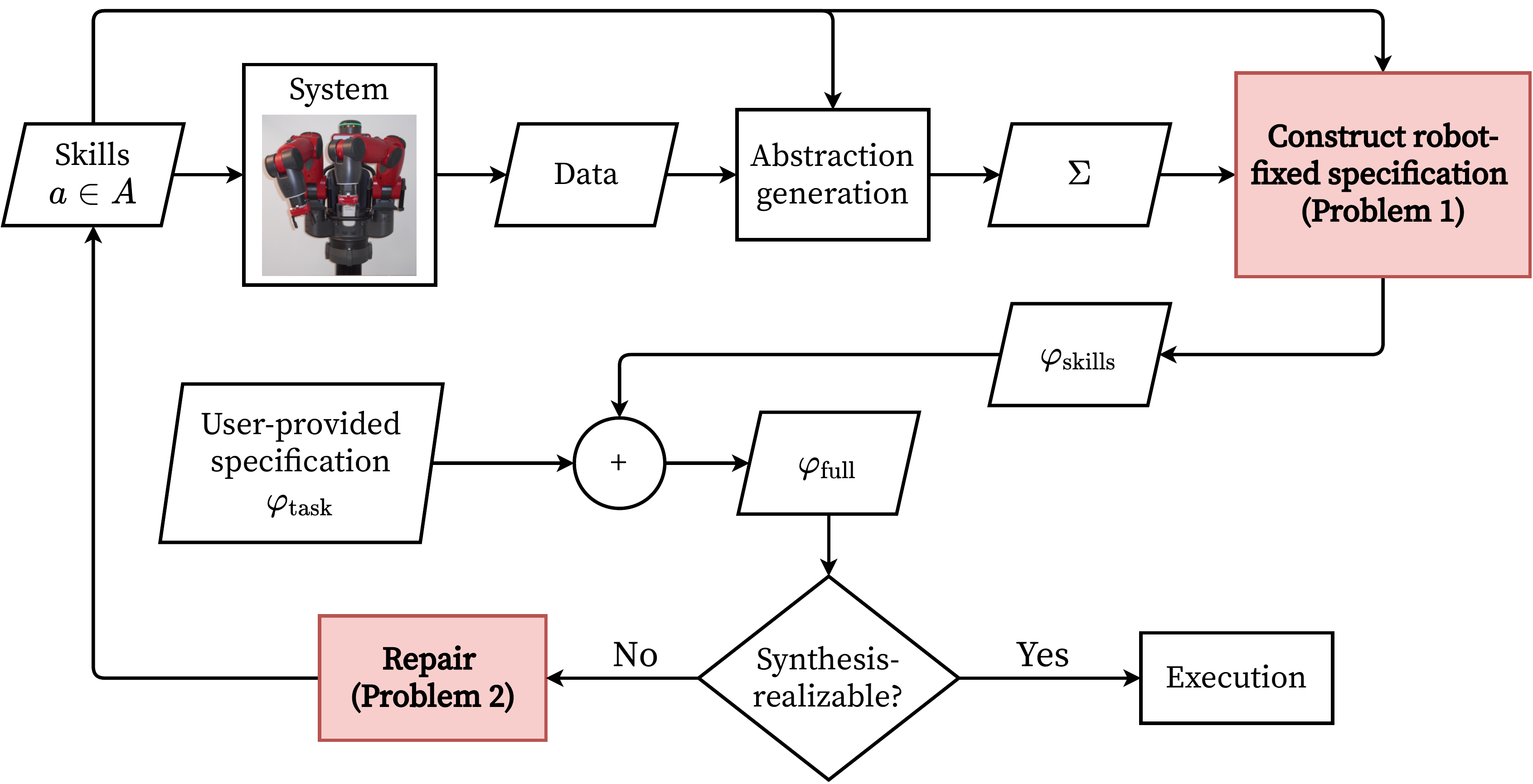}
\caption{Framework for automatically encoding robot capabilities, executing tasks, and repairing unrealizable tasks. Novel contributions are in bold and red. Figure adapted from \citet{pacheck2019automatic}.
}
\label{fig:overview}
\end{figure*}

Let $\vars$ be a set of atomic propositions and $\pi \in \vars$ be a Boolean variable. 
The syntax of a formula in \gls{ltl} \citep{pnueli1977temporal} obeys the following grammar:
\begin{equation*}
\varphi\ \Coloneqq\  \pi\ |\ \lnot \varphi\ |\ \varphi \vee \varphi\ |\ \bigcirc\varphi\ |\ \varphi\ \mathcal{U}\ \varphi
\end{equation*} 
where negation ($\lnot$,``not'') and disjunction ($\vee$,``or'') are Boolean operators and $\bigcirc$ (``next'') and $\mathcal{U}$ (``until'') are temporal operators. 
We define $\true = \varphi \vee \lnot \varphi$ and $\false = \lnot \true$. 
Given these operators, one can derive conjunction ($\varphi_1 \wedge \varphi_2 \equiv \lnot(\lnot\varphi_1\vee\lnot\varphi_2)$), implication ($\varphi_1\rightarrow\varphi_2 \equiv \lnot\varphi_1 \vee \varphi_2$), equivalence ($\varphi_1\leftrightarrow\varphi_2 \equiv (\varphi_1\rightarrow\varphi_2)\wedge(\varphi_2\rightarrow\varphi_1)$), eventually ($\lozenge\varphi\equiv \true\;\mathcal{U}\varphi$), and always ($\square\varphi\equiv\lnot\lozenge\lnot\varphi$). 
We define a \emph{symbolic state} as the set of all propositions that are currently $\true$ and denote all possible symbolic states by $\dom_\vars = 2^\vars$.
We use $\vars'=\set{\pi' \setor \pi \in \vars}$ as the set of primed versions of the variables in $\vars$ to denote variables at the next time step.
The set of all possible symbolic states at the next time step is $\dom_{\vars'} = 2^{\vars'}$.

The semantics of an \gls{ltl} formula $\varphi$ are defined over an infinite sequence $w = w_1 w_2 \ldots$~\citep{pnueli1977temporal}.
Each $w_i$ corresponds to the set of $\pi$ that are $\true$ at step $i$. 
We denote that a sequence $w$ satisfies an \gls{ltl} formula at instance $i$ by $w,i \models \varphi$. 
Intuitively, $w,i \models \bigcirc\varphi$ if $\varphi$ is $\true$ at step $i+1$, $w,i \models \square\varphi$ if $\varphi$ holds at every step after and including $i$ in $w$, and $w,i \models \lozenge\varphi$ if $\varphi$ holds at some step on or after $i$ in $w$. 

We consider the \gls{gr1} fragment of \gls{ltl}~\citep{bloem2012synthesis}. 
Let $\vars = \inp \cup \out$ be the set of atomic propositions, where $\inp$ is the state of the world as represented by the learned symbols $\Sigma$ and additional user-defined symbols $\user$, and $\out$ refers to the activation of robot skills, $\skills$. 
In \gls{gr1}, formulas are of the form:
\begin{equation}\label{eq:gr1_formula}
\begin{split}
\varphi &= \envform \rightarrow \sysform \\
\envform &= \varphi_i^e \wedge \varphi_t^e \wedge \varphi_g^e \\
\sysform &= \varphi_i^s \wedge \varphi_t^s \wedge \varphi_g^s
\end{split}
\end{equation}
where $\envform$ are assumptions about the environment's behavior and $\sysform$ are guarantees for the robot, also referred to as the system, and: 
\begin{flushitemize}
\item $\envinit$ and  $\sysinit$ are predicates over $\inp$ and $\inp \cup \out$, respectively, characterizing the initial states.
\item $\varphi^{e}_t$ and $\varphi^{s}_t$ are safety constraints of the form $\bigwedge_i \square \psi_i$ where $\psi_i$ are over $v$ and $\bigcirc u$ where $v \in \inp \cup \out$ for $\varphi^{e}_t$ and $\varphi^{s}_t$, and $u \in \inp$ for $\varphi^{e}_t$ and $u \in \inp \cup \out$ for $\varphi^{s}_t$.
\item $\envlive$ and $\syslive$ are the liveness requirements and characterize events that should occur infinitely often. 
Here, $\envlive = \bigwedge_{i=1}^{m} \square \lozenge J^e_i$ and $\syslive = \bigwedge_{j=1}^{n} \square \lozenge J^s_j$ where  $J^e_i$ and $J^s_j$ are predicates over $\inp \cup \out$.
\end{flushitemize}
An implementation of the specification is guaranteed to satisfy $\sysform$, provided that the environment satisfies $\envform$.


\subsection{Synthesis}

We use GR(1) synthesis~\citep{bloem2012synthesis} to find a strategy to accomplish a task.
In this work, a task consists of a set of system liveness guarantees ($\syslive$), initial conditions ($\envinit \wedge \sysinit$), and a set of ``hard'' system safety guarantees ($\syshard$).
When synthesizing, we consider two-player games played between a system and its environment where the system reacts to the environment~\citep{bloem2012synthesis}.
The environment is considered to be adversarial and attempts to keep the system from accomplishing its task.
This ensures the system is able to accomplish its task regardless of what happens in the environment.
We define a \emph{game structure} $\game = (\vars, \inp, \out, \initstates, \tenv, \tsys, \tsyshard, \obj)$ 
where $\vars$, $\inp$, and $\out$ are as defined in Section~\ref{sec:ltl}.
We define $\initstates$ as the set of states that satisfy $\envinit \wedge \sysinit$.
We define $\tenv \subseteq \dom_\vars \times \dom_{\inp'}$ as the set of current and next states satisfying $\envsafety$, 
$\tsys \subseteq \dom_\vars \times \dom_{\vars'}$ as the set of current and next states satisfying $\syssafety$, and $\tsyshard\subseteq \dom_\vars \times \dom_{\vars'}$ as the set of current and next states satisfying the hard system constraints in $\syshard$.
Hard system constraints in $\tsyshard$ cannot be modified by the synthesis-based repair process in Section~\ref{sec:synthesis_base_repair}.
Note that in Bloem,~et~al.~\citep{bloem2012synthesis}, $\tenv$ and $\tsys$ are defined as logical formulas; here we define them as sets of states.
The winning condition is given by $\obj = \envlive \rightarrow \syslive$.

Given a game structure $\game$, the \emph{realizability} problem is to decide if the game is winning for the system; either (a) for every environment action, the system is able to achieve $\varphi_s$ or (b) the system is able to falsify $\varphi_e$.
To determine if a specification is realizable, we find all the states $Z$ from which the system is able to win via a fixed point computation \citep{bloem2012synthesis}.
We iterate through every system liveness guarantee, $J^s_j$, and determine the set of states the system can always either transition to the next liveness goal from or falsify $\varphi_e$.
The \emph{synthesis} problem is to compute a strategy for the system to make the specification realizable~\citep{bloem2012synthesis}. 

We define a strategy computed using the synthesis process as $\strat = (\inp,\out,Q,Q_0,\delta,L)$, where:
\begin{flushitemize}
\item $\inp$ and $\out$ are the environment and system propositions, respectively, defined above
\item $Q$ is a set of states
\item $Q_0 \subseteq Q$ is the set of initial states
\item $\delta: Q \times 2^\inp \rightarrow Q$ is the transition function
\item $L: Q \rightarrow 2^\inp \times 2^\out$ is a labeling function that returns the propositions in $\inp \cup \out$ that are $\true$ in state $q\in Q$
\end{flushitemize}
Here, $\delta$ depends on $\inp$ as the system reacts to the environment state.

If Equation~\eqref{eq:gr1_formula} is unrealizable, meaning that there does not exist a strategy $\strat$ that will satisfy the task, the synthesis algorithm can provide a \emph{counter-strategy} that represents the behavior of the environment that will cause the system to fail to accomplish its task \citep{konighofer2009debugging,chatterjee2008environment}. 
We define a counter-strategy as $\counterstrat=(\inp,\out,Q,Q_0,Q_\textrm{n.o.t},\delta_\textrm{c.s.},L_\textrm{t},L_\textrm{n.o.t})$, where $\inp,\out,Q,Q_0$ are the same as in $\strat$ and:
\begin{flushitemize}
\item $Q_\textrm{n.o.t} \subseteq Q$ is the set of states from which there are no outgoing transitions (\textrm{n.o.t.})
\item $\delta_\textrm{c.s.}: Q\setminus Q_\textrm{n.o.t} \times 2^\inp \rightarrow Q $ is the transition function
\item $L_\textrm{t}: Q \setminus Q_\textrm{n.o.t} \rightarrow 2^\inp \times 2^\out$ is the labeling function for states with outgoing transitions
\item $L_\textrm{n.o.t}: Q_\textrm{n.o.t} \rightarrow 2^\inp$ is the labeling function for states with no outgoing transitions. The system has no valid transitions from $Q_\textrm{n.o.t}$, so only $\inp$ is needed to label $Q_\textrm{n.o.t}$.
\end{flushitemize}
In Section \ref{sec:enumeration_based_repair}, we use the states with no outgoing transitions, $Q_\textrm{n.o.t}$, to narrow the search for skills to repair unrealizable specifications in the enumeration-based repair approach. 

\section{Problem Formulation}\label{sec:pf}

Our goal is to automatically encode the capabilities of a robot in a \gls{ltl} formula and find a strategy for a reactive high-level task. 
If no strategy can be found, we find additional skills or modifications to skills that would allow the robot to complete the given task.

\textbf{Problem 1:} Given a set of skills $\skills$, automatically abstract and encode the capabilities of the robot in an \gls{ltl} formula, $\phifixed$. 
Allow a user to specify a reactive high-level task and find a strategy to fulfill it.

\textbf{Problem 2:} Given an unrealizable specification $\varphi_\textrm{unreal}$, find skill suggestions $\skills_\textrm{new}$, in the form of additional skills or modifications to current skills, such that constructing $\phifixed$ with $\skills \cup \skills_\textrm{new}$ makes the specification $\varphi_\textrm{unreal}$ realizable.

\section{Specification Encoding}

To address Problem 1, we automatically encode the robot's capabilities in $\phifixed$ using the symbols in $\learned$, which are learned from low-level sensor information \citep{konidaris2018skills}, and the skills $\skills$ of the robot.
The skills-based specification, $\phifixed$, can be reused for different tasks performed by the same robot.
The user then writes the task specific specification, $\phitask$, over $\learned \cup \user \cup \skills$, which is combined with $\phifixed$ to create $\phifull$. 
The set $\user$ contains additional user-defined environment propositions which correspond to signals the user wants the robot to react to.
We use a synthesis tool, such as Slugs \citep{ehlers2016slugs}, to either find a strategy, $\strat$, for accomplishing $\phifull$ if the specification is realizable or a counter-strategy, $\counterstrat$, if the specification is unrealizable. 
An overview of the framework is depicted in Figure \ref{fig:overview}.

\begin{figure*}[t]
\begin{equation}\label{eq:t_pre}
\syspre = \bigwedge_{a \in \skills} \square \left[ \lnot \left(\bigvee_{\sigma_p \in \precondsym} \left( \bigwedge_{\sigma \in \sigma_p} \bigcirc \sigma \right)\right)\rightarrow \lnot \bigcirc a\right]
\end{equation}
\begin{equation}\label{eq:sigma_false}
    \effectsymfalse{a}{j} = \bigcup_{f_q \textrm{ s.t. }\forall x_i \in f_q, \effmask{a}{j}(i)=\true}\{ \sigma \in \Sigma_{f_q} |\ \mathcal{G}(\sigma) \cap \mathcal{G}(\sigma_{a,j,f_q}) = \varnothing\}
\end{equation}
\begin{equation}\label{eq:env_trans}
\enveff = \bigwedge_{a \in \skills} \square \ \left[ a \rightarrow
 \bigvee_{j \in \{1,\ldots,k(a)\}} \left( \left( \bigwedge_{\sigma \in \effectsym{a}{j}}\bigcirc \sigma \right) \bigwedge \left( \bigwedge_{\sigma \in \effectsymfalse{a}{j}}\lnot \bigcirc \sigma \right) \bigwedge \left(\bigwedge_{\sigma \in \effectsymstay{a}{j}} (\sigma \leftrightarrow \bigcirc \sigma)\right)\right)\right]
\end{equation}
\end{figure*}


\subsection{Skills-Based Specification ($\phifixed$)}\label{sec:constant_spec}

The skills-based specification encodes the preconditions and postconditions of skills, along with mutual exclusion constraints on the skills and symbols.

Given a set of skills $\skills$, we first create symbols $\sigma \in \learned$, representing the effects of $a \in \skills$ \citep{konidaris2018skills}. 
We slightly abuse notation and use $a$ as a proposition that is $\true$ when the skill $a$ is active, and $\false$ otherwise. 

The skills-based specification ($\phifixed$) is composed of the system safety ($\syssafetyfixed = \syspre \wedge \sysmxskills$) and environment safety ($\envsafetyfixed = \enveff \wedge \envnoact \wedge \envmxsyms$) specifications. 
The system safety specification includes constraints on when the system is allowed to perform skills ($\syspre$) and optionally the mutual exclusion of skills ($\sysmxskills$).
The environment safety specification includes how each $\sigma$ is allowed to change with the application of a skill $(\enveff)$, the effect of no skill being performed ($\envnoact$), and the mutual exclusion of symbols over the same factor ($\envmxsyms$). 

\subsubsection{System Safety ($\syssafetyfixed$):}
We encode constraints on when skills can be performed in $\syspre$ based on the preconditions of the skills.
For each action, we find all possible combinations of symbols that overlap with the precondition mask and determine which combinations fall within the precondition set \citep{konidaris2018skills}. 
We define $\precondsym = \{\sigma_p \in \sympremask | \mathcal{G}(\sigma_p) \subseteq \precondset\}$, where $\sympremask = \prod_{f_q\in F\textrm{ s.t. }\forall x_i\in f_q, \premask[a](i)=\true} \Sigma_{f_q}$. 
The set $\precondsym$ contains all the combinations of $\sigma$ that satisfy the precondition of $a$. 
We encode in $\syspre$ that when none of the preconditions in $\precondsym$ are satisfied, the robot is not allowed to perform $a$ as shown in Equation~\eqref{eq:t_pre}.
Equation~\eqref{eq:t_pre} states that skill $a$ cannot be executed at the next step when no combinations of symbols $\sigma_p \in \precondsym$ are $\true$ at the next step.
This allows the robot to choose to execute a skill only when the preconditions of a skill are satisfied.
We write $\syspre$ over $\dom_{\vars'}$ instead of over $\dom_\vars$ as in \citet{pacheck2019automatic} to generate additional types of skill suggestions in Section~\ref{sec:synthesis_base_repair} and match assumptions made in \citet{pacheck2020finding}.
In Figure~\ref{fig:sym_gen}, $\precondsym[a_2] = \{\{\sigma_{a_1,1,x_1}\},\allowbreak\{\sigma_{a_1,2,x_1}\}\}$.

We can encode mutual exclusion of skills in $\sysmxskills$ at both the current and next step.
In the examples presented, skills are mutually exclusive, although in general they need not be.

Note that $\sysmxskills$ is considered a ``hard'' constraint ($\syshard$) for synthesis-based repair (Section~\ref{sec:synthesis_base_repair}) and so is not allowed to be changed.
On the other hand, $\syspre$ is not a ``hard'' constraint and can be modified during synthesis-based repair, meaning that we can modify the preconditions of the skills.

\subsubsection{Environment Safety ($\envsafetyfixed$):} 
To encode a skill's (possibly nondeterministic) effects, we consider the skill outcome to be determined by the environment.

We denote the symbols which become $\true$ with the application of a skill $a$ as $\effectsym{a}{j} = \cup_{f_q \textrm{ s.t. }\forall x_i \in f_q, \effmask{a}{j}(i)=\true}\ \sigma_{a,j,f_q}$\citep{konidaris2018skills}. 
In Figure \ref{fig:sym_gen}, $\effectsym{a_1}{2} = \{\sigma_{a_1,2,x_1}\}$.

When $a$ is applied, symbols belonging to the same factor $f_q$ whose grounding sets do not overlap with those in $\effectsym{a}{j}$ become $\false$ due to mutual exclusion. 
We denote this set of symbols $\effectsymfalse{a}{j}$ in Equation~\eqref{eq:sigma_false}.
In Figure \ref{fig:sym_gen}, $\effectsymfalse{a_1}{2} = \{\sigma_{a_1,1,x_1},\sigma_{a_2,1,x_1}\}$.

When performing synthesis \citep{kress2018synthesis}, if a symbol is not constrained, it can be set to any value. 
We must therefore consider the ``frame problem'' \citep{ghallab2004automated} and constrain symbols that are not modified by the current skill to stay the same. 
The set $\effectsymstay{a}{j} = \cup_{f_q\textrm{ s.t. }\forall x_i \in f_q,\textrm{eff-mask}^j(a)(i)=\false}\Sigma_{f_q}$ contains the $\sigma$ not modified by skill $a$ in the $j^{th}$ outcome. 
In Figure \ref{fig:sym_gen}, because $x_2$ is not modified in effect 2 of $a_1$, $\effectsymstay{a_1}{2} = \{\sigma_{a_1,1,x_2},\sigma_{a_2,1,x_2}\}$.

We encode how the truth values for $\sigma$ can change when a skill is applied in $\enveff$ in Equation~\eqref{eq:env_trans}.
Equation \eqref{eq:env_trans} states that when skill $a$ is performed, it leads to one of $j$ nondeterministic outcomes with $\sigma \in \effectsym{a}{j}$ becoming $\true$, $\sigma \in \effectsymfalse{a}{j}$ becoming $\false$, and the truth value of $ \sigma \in \effectsymstay{a}{j}$ remaining the same.
Symbols whose grounding sets overlap with those in $\effectsym{a}{j}$ and are therefore not in $\effectsym{a}{j}$, $\effectsymfalse{a}{j}$, or $\effectsymstay{a}{j}$ are not constrained. 
In the examples presented in this work, there are no symbols whose grounding sets overlap that have not been merged into one symbol.
During the synthesis process, the adversarial environment chooses which nondeterministic outcome $j$ would result in the worst case scenario for the system.
This enables us to guarantee that no matter what the nondeterministic effect of an action is, the system is still able to complete its task.

When no skill is performed, we encode in $\envnoact$ that the truth values of $\sigma$ remain the same.
\begin{equation}
    \begin{split}
        \envnoact = \square \left[ \left( \bigwedge_{a \in \skills} \lnot a \right) \rightarrow \left( \bigwedge_{\sigma \in \Sigma} (\sigma \leftrightarrow \bigcirc \sigma)\right)\right]
    \end{split}
\end{equation}

We encode the mutual exclusion of non-overlapping symbols over the same factor in $\envmxsyms$ at both the current and next step.
We enforce that only one of the symbols in a factor is $\true$ at a time.
In Figure~\ref{fig:sym_gen}, $\sigma_{a_1,1,x_1}$, $\sigma_{a_1,2,x_1}$, and $\sigma_{a_2,1,x_1}$ are all grounded over $x_1$ and do not overlap, so only one of them can be $\true$ at a time.


\subsection{Task Specification, Synthesis, and Execution}
\label{sec:taskspec}

The user writes the task-specific specification, $\phitask$, which may include additional environment propositions $\varuser \in \user$.
The task-specific specification can include constraints on the initial state(s) of the system and environment, system liveness, and environment liveness in $\sysinittask$, $\envinittask$, $\syslivetask$, and $\envlivetask$, respectively.
Additional system safety constraints are added in $\syssafetytask$, which we consider to be a ``hard'' constraint and which is not allowed to be changed during the synthesis-based repair. 
Tasks can encode objectives such as repeatedly  accomplishing a goal or goals, always avoiding some states, always making sure a constraint holds, or reacting to environment events.
We give examples of tasks and $\syslivetask$, $\envlivetask$, and $\syssafetytask$ in Section \ref{sec:demos}.

The full specification $\phifull$ is shown in Equation~\eqref{eq:full_spec}.
We generate a strategy for satisfying $\phifull$ using a synthesis tool, such as Slugs \citep{ehlers2016slugs}. 
If $\phifull$ is realizable, the resulting strategy $\strat = (\inp,\out,Q,Q_0,\delta,L)$, where $\inp = \learned \cup \user$ and $\out = \skills$, is used to control the robot.
If $\phifull$ is not realizable, we repair the specification using either an enumeration-based or synthesis-based repair approach (Section~\ref{sec:repair}).

\begin{figure*}
    \begin{equation}\label{eq:full_spec}
        \phifull = \envinittask \wedge \overbrace{\enveff \wedge \envnoact \wedge \envmxsyms}^{\envsafetyfixed} \wedge \envlivetask \rightarrow
        \sysinittask \wedge \rlap{$ \overbrace{\phantom{\syspre \wedge \sysmxskills}}^{\syssafetyfixed}$} \syspre \wedge \underbrace{\sysmxskills \wedge \syssafetytask}_{\syshard} \wedge \syslivetask
    \end{equation}
\end{figure*}

To assist the user in writing $\phitask$, we visualize the grounding of the symbols and combinations of symbols.
Figures \ref{fig:robot_formula}(E,F) and \ref{fig:precondition_effect_plates} show examples of individual symbol groundings. 
Figures~\ref{fig:robot_formula}(A-D), \ref{fig:blocks_liveness}(C,F), \ref{fig:plates_liveness}(A-D), \ref{fig:vials_liveness}(B,D), \ref{fig:plates_sequence}, \ref{fig:synthesis_suggestions_unrealizable_1_and_2}, \ref{fig:synthesis_suggestions_unrealizable_3_and_4}, \ref{fig:plate_fault_model}, and \ref{fig:vaccine_fault_model}(G-I) visualize the combination of multiple symbols. 
To visualize each combination of symbols, we sample from the intersection of the grounding sets of the symbols.

\section{Specification Repair}\label{sec:repair}


We address Problem 2 of making an unrealizable specification realizable by searching for additional skills or modifications to existing skills.
We present and compare both an enumeration-based and synthesis-based approach, based on methods first proposed in \citet{pacheck2019automatic} and \citet{pacheck2020finding}, respectively.

\subsection{Enumeration-Based Repair}\label{sec:enumeration_based_repair}

In the enumeration-based repair approach, we search for one or more skills, $\anew \in \skills_\textrm{new}$, that would make an unrealizable task realizable when $\phifixed$ is constructed with $\skills \cup \skills_\textrm{new}$.
We build on the enumeration-based repair process presented in \citet{pacheck2019automatic}.
There, we assume that only one skill, $\anew$ is required to repair the specification.
In this work, we relax that assumption and repair specifications that may need more than one additional skill.
We assume the robot has all the symbols it needs to define the task.
We also assume that our new skills will consist of a precondition set and effect mask we have already seen, restricting the search space for the new skills.
By assuming our new skills will consist of a precondition set and effect mask we have already seen, we will not find all possible skills to repair the task and may even be unable to repair the specification.
It is possible to relax these assumptions to consider all possible preconditions and postconditions; however, without these assumptions, the number of possible skills is too large to reasonably consider.
In this work we are able to find skill suggestions for all examples in Section~\ref{sec:demos} with the enumeration-based approach while making the above assumptions.

We leverage the structure of $\counterstrat$ to focus the repair process. 
The counter strategy, $\counterstrat$, contains the environment behaviors that make a specification unrealizable. 
In general, a \gls{gr1} specification is unrealizable either because (i) the robot violates safety constraints, (ii) gets stuck in a loop when trying to satisfy its liveness goals, or (iii) is unable to reach the liveness goals from its initial conditions. 
When the robot can only satisfy at least one of its liveness goals by using skills that leave the environment unable to act, the counter strategy contains states with no successors (i.e. $Q_\textrm{n.o.t.} \neq \varnothing$). 
We find the skills that lead to these states, and use their precondition sets to narrow the search space for $\skills_\textrm{new}$.
Then, we generate new effect sets, based on existing effect masks, and combine them with existing precondition sets to create new skills.

Algorithm~\ref{alg:repair1} shows our enumeration-based procedure for repairing unrealizable specifications. 
On Line~\ref{eq:new_eff}, we create new effect sets, $\Sigma^{+}$, based on existing effect masks, based on the assumption that new skills will change similar states as current skills. 
For each existing effect mask, we find all the state variables that are in the mask. 
We then compute all possible combinations of $\sigma_{a,j,f_q}$ that ground to those state variables, regardless of which skill they were originally generated from.

On Line~\ref{eq:find_presyms}, we find $\skills_\textrm{n.o.t.}$: the set of skills whose preconditions were satisfied that lead to states with no outgoing transitions.
Based on $\skills_\textrm{n.o.t.}$, we then construct a set of candidate skills, $\skills_\textrm{candidate-skills-n.o.t}$ on Line~\ref{eq:new_skills_test}.
Each one consists of the precondition set of a skill in $\skills_\textrm{n.o.t.}$ and a new effect set found in $\Sigma^{+}$.
We then construct a second set of new skills, $\skills_\textrm{candidate-skills-all}$, in Line~\ref{eq:new_skills_all} based on the precondition sets of all skills $\skills$.
Slightly abusing notation, we denote candidate skills in $\skills_\textrm{candidate-skills-n.o.t}$ and $\skills_\textrm{candidate-skills-all}$ as pairs containing the preconditions of a skill and which symbols become $\true$, while using the name of a skill to denote skills in $\skills_\textrm{n.o.t}$.

We then consider combinations of a skill in $\skills_\textrm{candidate-skills-n.o.t}$ and $n_\textrm{new-skills-desired} - 1$ skills in $\skills_\textrm{candidate-skills-all}$.
We assume that one new skill needs to include a precondition from the skills $\skills_\textrm{n.o.t.}$, but do not assume any other new skills need to start from one such precondition.
In Lines~\ref{eq:loop_skills}-\ref{eq:loop_end}, we write the unrealizable specification with the new skills and attempt to synthesize a strategy.
If the specification is realizable, we store the skill combination.
All $\skills_\textrm{new}$ which make $\phifull$ realizable are returned to the user, enabling them to select the skill they deem easiest to physically implement.


\begin{algorithm*}[t]
\SetAlgoLined
\DontPrintSemicolon
\KwIn{$\counterstrat(\inp,\out,Q,Q_0,Q_\textrm{n.o.t},\delta_\textrm{c.s.},L_\textrm{t},L_\textrm{n.o.t}),n_\textrm{new-skills}, \skills, \phitask$}
\KwOut{$\alpha_\textrm{suggestions}$}
$\Sigma^+ :=  \bigcup_{a\in \skills,j\in\{1,\ldots,k(a)\}} \prod\nolimits_{f_q \in \factors \textrm{s.t.} \forall x_i \in f_q, \effmask{a}{j}(i)=\true} \Sigma_{f_q}$\tcp*[r]{Create new postconditions} \label{eq:new_eff}
$\skills_\textrm{n.o.t} := \{a \in \skills |\ \exists \sigma_p \in \precondsym[a],\ q\in Q,\ e \in 2^{\inp} \textrm{s.t.}\ \delta(q,e) \in Q_\textrm{n.o.t},\ \sigma_p \in L_\textrm{t}(q)\}$ \tcp*{Skills leading to states with no outgoing transitions}\label{eq:find_presyms}
$\skills_\textrm{candidate-skills-n.o.t} := \{(\sigma_{\textrm{pre}(a_\textrm{n.o.t})}, \sigma) \setor a_\textrm{n.o.t} \in \skills_\textrm{n.o.t}, \sigma \in \Sigma^{+} \textrm{ s.t. } \nexists\ a_\textrm{orig} \in \skills \textrm{ where } \sigma_{\textrm{pre}(a_\textrm{n.o.t})} = \sigma_{\textrm{pre}(a_\textrm{orig})} \textrm{ and } \sigma_{\textrm{eff}(a_\textrm{n.o.t})}^{\true} = \sigma_{\textrm{eff}(a_\textrm{orig})}^{\true} \}$\tcp*{New skills based on $\skills_{n.o.t}$}\label{eq:new_skills_test}
$\skills_\textrm{candidate-skills-all} := \{(\sigma_{\textrm{pre}(a)}, \sigma) \setor a \in \skills, \sigma \in \Sigma^{+} \textrm{ s.t. } \nexists\ a_\textrm{orig} \in \skills \textrm{ where } \sigma_{\textrm{pre}(a)} = \sigma_{\textrm{pre}(a_\textrm{orig})} \textrm{ and } \sigma_{\textrm{eff}(a)}^{\true} = \sigma_{\textrm{eff}(a_\textrm{orig})}^{\true} \}$\tcp*{New skills based on $\skills$}\label{eq:new_skills_all}
$\alpha_\textrm{suggestions} := \varnothing$\;
$\alpha_\textrm{combinations}$ = combinations of $\skills_\textrm{candidate-skills-n.o.t}$ and $(n_\textrm{new-skills}-1)$ skills from $\skills_\textrm{candidate-skills-all}$\;\label{eq:combinations_alg}
\For{$\skills_\textrm{new} \in \alpha_\textrm{combinations}$\label{eq:loop_skills}}{
    Write $\phifixed$ with $\skills \cup \skills_\textrm{new}$\; \label{eq:write}
    Synthesize $\phifull\ \textrm{with}\ \phifixed\ \textrm{and}\ \phitask$\;\label{eq:synthesize}
    \If {Realizable}{
        $\alpha_\textrm{suggestions} := \alpha_\textrm{suggestions} \cup \skills_\textrm{new}$\; \label{eq:save_skill}
    }
}\label{eq:loop_end}
\textbf{return} $\alpha_\textrm{suggestions}$\;
\caption{Enumeration-Based Repair}\label{alg:repair1}
\end{algorithm*}

\subsection{Synthesis-Based Repair}\label{sec:synthesis_base_repair}


\begin{algorithm*}[t]
\KwIn{Game structure $\game$, Winning states $\winning$, User-defined variables $\user$, {\color{blue} Extra skills $\skills_\textrm{extra-skills}$, Extra skill that can be changed $a_\textrm{extra-skill-modify}$}} 
\KwOut{Updated $\tenv$}
$R_1:=\{(\varstate, \varinp')\setor \varstate \in \dom_\vars, \varinp \in \dom_\inp, 
\exists \varout \in \dom_\out\ \textrm{s.t.}\ (\varstate, \varinp', \varout') \in \tsys\ \textrm{and}\ (\varinp,\varout) \in Z\}$\; \label{restrictPost:existSys}
$R_2:=\{\varstate \in \dom_\vars \setor \exists \varinp \in \dom_\inp\ \textrm{s.t.}~ (\varstate,\varinp')\in \tenv\ \textrm{and}\ (\varstate,\varinp')\in R_1\}$\;\label{restrictPost:existEnv}
$R_2:= R_2 \backslash \winning$\; \label{restrictPost:removeAlreadyWinning}
${\color{blue}R_2 := \{(v_e, v_s) \in R_2 \setor \textrm{ if } v_s \in \skills_\textrm{extra}, v_s = a_\textrm{extra-skill-modify}\}}$\;\label{restrictPost:onlyOneSkill}
$\tenvnew:= \{(\varstate,\varinp') \setor \varstate\in \dom_\vars, \varinp \in \dom_\inp \ \textrm{s.t.} \ \varstate \in R_2 \ \textrm{and}\ (\varstate,\varinp')\in R_1 \cap \tenv\}$\;\label{restrictPost:newTransitions}
${\color{blue} \tau_e^\textrm{new-expanded} = \{ (v_{\vars\textrm{-expanded}}, v_{e\textrm{-expanded}}') \setor \exists  (v_\vars, v_e') \in \tenvnew, r_1, r_2 \in 2^\user \textrm{ s.t. }}$ ${\color{blue} (v_{\vars\textrm{-expanded}} = v_\vars \setminus r_1 \textrm{ or } v_{\vars\textrm{-expanded}} = v_\vars \cup r_1) \textrm{ and } (v_{e\textrm{-expanded}}' = v_e' \setminus r_2' \textrm{ or } v_{e\textrm{-expanded}}'=v_e' \cup r_2')\}}$\;\label{restrictPost:new_alg_line}
$\tau_e^{old} := \{(\varstate,\varinp') \setor \varstate\in \dom_\vars, \varinp\in\dom_\inp \ \textrm{s.t.} \ \varstate\notin R_2 \ \textrm{and}\ (\varstate,\varinp')\in \tenv\}$\;\label{restrictPost:oldTransitions}
return $\tau_e^\textrm{new-expanded} \cup \tau_e^{old}$\;\label{restrictPost:allTransitions}
\caption{\textbf{restrictPostconditions} (additions to \citet{pacheck2020finding} are in {\color{blue} blue})}
\label{alg:restrictPost}
\end{algorithm*}

Our second approach to repair unrealizable specification is synthesis-based repair where, as opposed to the enumeration-based approach, we take advantage of the synthesis process to guide the repair.
We extend the synthesis-based repair introduced in \citet{pacheck2020finding} to find suggestions for skills that can repair reactive tasks.
Here we give a brief overview of the process (see \citet{pacheck2020finding} for a full description), and describe modifications we have made to the repair process that allow us to find suggestions for a larger class of specifications.
With these modifications we are able to find repair suggestions for specifications with reactive liveness guarantees, which is not possible with the repair process in \citet{pacheck2020finding}.

The repair process takes an unrealizable specification and finds suggestions of new skills or modifications to existing skills.
Modifications to existing skills are in the form of additional preconditions that should be added to skills, thereby allowing the robot to use them in additional situations, or postconditions that should be removed from skills, essentially reducing nondeterminism.
Additional preconditions are a set of symbol combinations $\sigma_{\textrm{pre}(a)}^{\textrm{added}}$ that should be added to the existing preconditions, i.e. $\sigma_{\textrm{pre}(a)}^{\textrm{new}} = \sigma_{\textrm{pre}(a)} \cup \sigma_{\textrm{pre}(a)}^{\textrm{added}}$.
The repair process can also remove one or more postconditions $j\in\{1,\ldots,k(a)\}$.

To find new skills, we add a set of additional skills, $\skills_\textrm{extra-skills}$, to the specification that are unrestricted---they can be executed from any combination of symbols and can result in any combination of symbols.
The preconditions of $a_\textrm{extra-skill} \in \skills_\textrm{extra-skills}$ are $\precondsym[a_\textrm{extra-skill}] = 2^\learned$.
In practice, we simply do not include constraints on the preconditions of $\skills_\textrm{extra-skills}$ in $\syssafetyfixed$.
Similarly, there are no constraints on the postconditions of $a_\textrm{extra-skill} \in \skills_\textrm{extra-skills}$.
The extra skills still need to satisfy constraints on the mutual exclusion of symbols in $\envmxsyms$.
New skills are of the form of a set of preconditions, $\sigma_{\textrm{pre}(a)}^{\textrm{new}}$ and collection of postconditions for that skill $\sigma_{\textrm{eff}^j(a)}^{\true-\textrm{new}}$.

To find skill suggestions, the synthesis-based repair process performs synthesis until it determines the specification is unrealizable.
The repair process then iteratively modifies $\tenv$ and $\tsys$, which correspond to $\envsafetyfixed$ and $\syssafetyfixed$, respectively, and attempts to perform synthesis until the specification is realizable.
The repair process returns $\tenvnew$ and $\tsysnew$, from which we extract the skill suggestions.

\begin{figure*}[t]
    \centering
    \includegraphics[width=\textwidth]{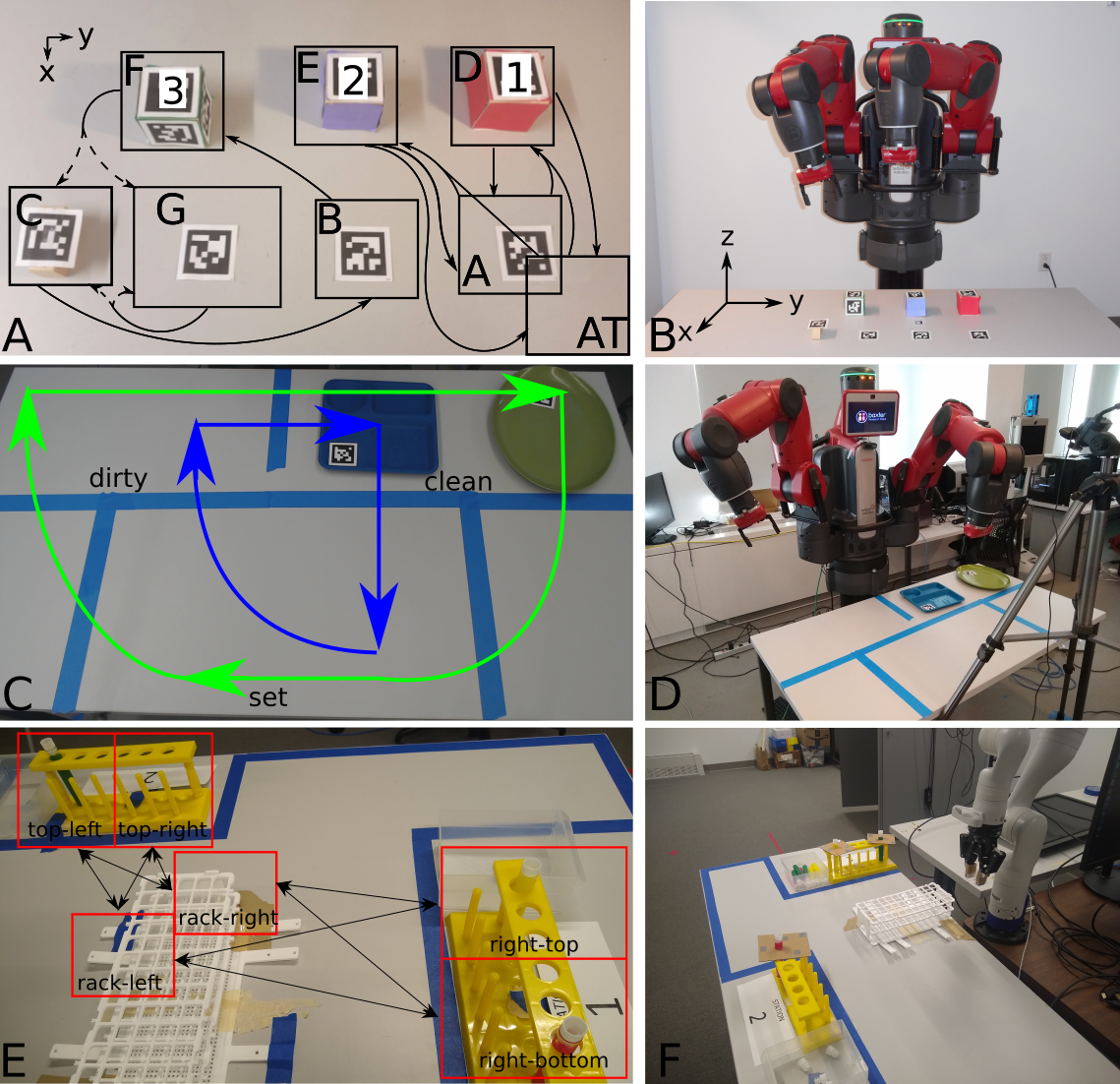}
    \caption{
    \textbf{Baxter Blocks}:
    (A) The arrows show the skills given to the Baxter robot to move blocks between lettered locations. 
    Dashed arrows represent skills with nondeterministic outcomes. 
    Note that one skill takes blocks from A and AT to D and another from A and AT to E while all the other skills have preconditions that are only one location. 
    (B) Initial setup of the Baxter Blocks example.
    \textbf{Baxter Plates}:
    (C) The colored arrows represent the skills given to the Baxter robot to move the two plates between the clean, set, and dirty locations. 
    The skill moving the green (oval) plate from clean to set is not possible when the blue (square) plate is set and the skill moving the green plate from dirty to clean is not possible when the blue plate is dirty or clean.
    The skill moving the blue plate from set to dirty is not possible when the green plate is set. 
    (D) Initial setup of the Baxter Plates example. 
    \textbf{Kinova Vials}:
    (E) The arrows represent the skills given to the Kinova robot to move the vials. 
    The skills move the green, red, and yellow vials between the top-left, top-right, rack-left, rack-right, right-top, and right-bottom locations. 
    The skills move the vials between the yellow (top and right) and white (rack) holders (and vice-versa) but not between the two yellow holders. 
    (F) Initial setup of the Kinova Vials example.
    (A) and (B) are from \citet{pacheck2019automatic}.
    }
    \label{fig:env_setup}
\end{figure*}

To modify $\tenv$, the process of restricting postconditions takes the current game structure $\game$, a set of winning states $Z$, the user-defined variables $\user$, the set of extra skills $\skills_\textrm{extra-skills}$, and which extra skill can be changed on this iteration $a_\textrm{extra-skill-modify}$.
The process of restricting the postconditions removes transitions from $\tenv$ as shown in Algorithm~\ref{alg:restrictPost}.
We start with the current set of winning states $Z$ and attempt to expand it.
In Line~\ref{restrictPost:existSys}, we find $R_1$, the set of states from which the system has the ability to reach $Z$.
Then, in Line~\ref{restrictPost:existEnv}, we find $R_2$, the set of states from which at least one next state will be in $R_1$.
This set includes states where one nodeterministic action outcome may reach $R_1$ (and can therefore reach $Z$) but another may not.
We then remove already winning states from $R_2$ in Line~\ref{restrictPost:removeAlreadyWinning}.
In \citet{pacheck2020finding}, we find $\tenvnew$ by removing all the postconditions associated with $R_2$ that do not result in $R_1$ in Line~\ref{restrictPost:newTransitions} (skipping Line~\ref{restrictPost:onlyOneSkill}).
We then return $\tenvnew$ and $\tau_e^{old}$ (the transitions that were not modified).

In \citet{pacheck2020finding}, the repair process can remove any transition, while running \texttt{restrictPostconditions}, in order to attempt to repair the specification.
In the case of reactive specifications, this can result in unwanted behavior.
The process of restricting postconditions is able to remove transitions that correspond to changes in the values of $\varuser \in \user$, as symbols in both $\user$ and $\learned$ are treated the same.
As a result, the synthesis-based repair process proposed in \citep{pacheck2020finding} may offer suggestions for skills that can change (or keep the same) the value of symbols $\varuser \in \user$.
We show an example of this as mentioned in the Baxter Plates example in Section~\ref{sec:demos}.

In this work, we modify Algorithm~\ref{alg:restrictPost} to repair reactive specifications. 
Modifications to the algorithm are shown in blue.
We add Lines~\ref{restrictPost:onlyOneSkill} and \ref{restrictPost:new_alg_line} to Algorithm~\ref{alg:restrictPost}.
We do not allow the repair process to change the value of symbols $\varuser \in \user$.
The modifications in $\tenvnew$ found in Line~\ref{restrictPost:newTransitions} may include restrictions on the truth value of $\varuser \in \user$. Since we do not want to allow such modifications, 
for each possible modification in $\tenvnew$, we add transitions corresponding to all possible changes in the truth value of $\varuser \in \user$ in Line~\ref{restrictPost:new_alg_line}.
Now, the environment is able to make transitions with no restriction on $\varuser \in \user$, while still having restrictions on $\sigma \in \Sigma$.
The addition of Line~\ref{restrictPost:new_alg_line} does not allow us to find any suggestions for some specifications.
As we show in Section~\ref{sec:demos}, for some specifications, the only way for the synthesis-based repair process to find suggestions is by changing the values of variables $\varuser \in \user$, essentially trying to enforce the behavior of external events,  which is not desired.

For some specifications, multiple skills are required to find suggestions.
To account for this, we only allow the repair process to change one additional skill $a_\textrm{extra-skill-modify} \in \skills_\textrm{extra-skills}$ per iteration in (Line~\ref{restrictPost:onlyOneSkill}).
Without Line~\ref{restrictPost:onlyOneSkill}, the repair process attempts to change all of the extra skills to the same postconditions at once, which can cause the repair process to fail.

Additionally, we only perform one iteration of the \texttt{while} loop in Algorithm~2: Repair of \citet{pacheck2020finding}.
In the previous work, we modified $\tsys$ and $\tenv$, then applied the controllable predecessor operator to $Z$ with the new $\tsys$ and $\tenv$ until the current liveness guarantee overlapped with $Z$.
However, because we are making modifications but not allowing the user-defined variables to be restricted, the process of restricting the postconditions may expand $Z$ in ways not captured by only the application of the controllable predecessor.
The new skills may allow for the system to have more control over the outcome of skills at other states not currently in $Z$.
By performing synthesis after only one iteration of the repair process, we are able to find suggestions for more specifications.

We find $\tsysnew$ by relaxing the preconditions as in \citet{pacheck2020finding}.
The process is similar to restricting the postconditions.
We find all the states from which the system can reach $Z$ without violating $\tsyshard$.
We then find the states that will always lead to these states from which the system can reach $Z$.
We add these states $\tsys$ and thereby expand the preconditions.

After finding $\tenvnew$ and $\tsysnew$ that allow the specification to be synthesized, the repair process finds a  strategy for the system to achieve the liveness guarantees from the initial conditions.
We then compare the preconditions and postconditions of the skills performed during the strategy to the preconditions and postconditions of the skills initially given to the robot.
The new postconditions for the extra skills are then those seen in the strategy.
The new preconditions for the extra skills are those seen in the strategy.
Similarly, the additional preconditions for extra skills are those seen in the strategy \citep{pacheck2020finding}.

To find multiple suggestions, once one suggestion has been found by the synthesis-based repair process, the new skills making up the suggestion are disallowed along with any additional preconditions.
A new strategy is found if possible and another suggestion extracted.
We continue finding additional suggestions and disallowing previous suggestions until there are no more suggestions \citep{pacheck2020finding}.

\subsection{Enumeration-Based vs Synthesis-Based Repair}

We demonstrate the repair of unrealizable specifications using both the enumeration-based and synthesis-based approach in Section~\ref{sec:demos}.

\begin{table*}[]
\centering
\begin{tabular}{@{}c@{}s@{}s@{}s@{}s@{}s@{}}
\toprule
                & Number of given skills & Number of partitioned skills & Number of symbols generated & Number of pre/post pairs collected & Number of formulas in $\phifixed$ \\ \midrule
Baxter Blocks   & 9                      & 20                           & 19                          & 1052        & 83                     \\
Baxter Plates   & 6                      & 7                            & 9                           & 100         & 35                      \\
Kinova Vials & 48                     & 48                           & 18                          & 860            & 195                    \\ \bottomrule
\end{tabular}
\caption{Overview of the different robot demonstration environments. 
For each demonstration environment, we gave the robot a set of skills, collected data on the preconditions and postconditions, generated symbols, and automatically encoded the preconditions and postconditions in an \gls{ltl} formula. 
Partitioned skills are those that have different effects from different preconditions.
The number of formulas in $\phifixed$ include constraints on the preconditions, postconditions, mutual exclusion of skills, and mutual exclusion of symbols.
}
\label{tab:demo_overview}
\end{table*}

Table~\ref{tab:repair_overview} shows the synthesis-based repair is faster than the enumeration-based repair, especially as the number of skills and symbols increases.
The difference in time to find suggestions was especially apparent in the Kinova Vials example, where we were not able to run the enumeration-based repair to completion.
This disparity is pronounced because the enumeration-based repair needs to enumerate all possible combinations of skills, which does not scale well when there are multiple skills required to repair a specification.
It is possible to terminate both repair processes early and only receive a portion of the suggestions; however, it is not possible to know at which point in the repair process the suggestion desired by the user will be found.

While the enumeration-based repair takes longer than the synthesis-based repair, the suggestions returned tend to be more interpretable due to both the number and type of skills suggested.
The enumeration-based repair attempts to find skills that have the preconditions of existing skills and the effect masks of existing skills.
As a result, the suggested skills will look similar to the existing skills.
The synthesis-based repair process suggests skills with postconditions and preconditions that do not necessarily look similar to the existing skills.
There is no limit on the number of skills provided by the synthesis-based repair process in a single suggestion.
This can make suggestions more difficult to interpret.
For example, in the suggestion shown in Figure~\ref{fig:vaccine_fault_model}, the synthesis-based repair process suggests 3 skills, while the enumeration based repair process only suggestion 2 skills.
For the synthesis-based repair process, the skills that are suggested are highly dependent on the choices the system makes during the process of finding a strategy.
Changing the order of states visited by the system during the determinization process will likely result in different suggestions.

The synthesis-based repair does not always find suggestions to repair the specification.
As shown in Section~\ref{sec:demos}, there are certain specifications for which the only way the synthesis-based repair can provide suggestions is by suggesting skills that change the value of user-defined variables, essentially enforcing a behavior on uncontrolled events .

\section{Robot Demonstrations}\label{sec:demos}

We demonstrate automatically creating $\phifixed$, writing and executing task specifications, and the repair process with examples involving a Kinova robot manipulating vials and a Baxter robot manipulating blocks and pushing plates.
The Baxter Blocks example shows skills with nondeterminism and several unrealizable specifications.
The Baxter Plates example shows the use of raw camera images to create symbols and the benefit of enumeration-based repair over synthesis-based repair.
The Kinova Vials example shows the benefits of the synthesis-based repair approach over the enumeration-based repair approach.


\subsection{Environment Setup}

We have three different demonstration environments.

\textbf{Baxter Blocks:} In the Baxter Blocks example a Baxter robot is manipulating blocks on a table as shown in Figure~\ref{fig:env_setup}(A, B).
There are three blocks: red, blue and green (also labelled 1, 2, and 3, respectively).
The red and blue blocks can be placed at locations A, D and E, and can also be stacked at AT.
The green block can be placed at locations B, C, F, and G.
The location of the blocks are determined by AprilTags~\citep{Wang2016apriltags} detected by the Baxter's wrist cameras.
The state space is the $x$, $y$, and $z$ position of each of the blocks.

\textbf{Baxter Plates:} The Baxter Plates example contains a Baxter robot manipulating plates on a table as shown in Figure~\ref{fig:env_setup}(C, D).
There are two plates (a blue square plate and a green oval plate) that can be moved between the clean, set, and dirty regions.
The state in this example is the image recorded by the USB camera shown in the right of Figure~\ref{fig:env_setup}D.

\textbf{Kinova Vials:} The Kinova Vials example deals with a Kinova arm moving three colored vials (green, red and yellow) as shown in Figure~\ref{fig:env_setup}(E, F).
The vials can be in six regions: top-left, top-right, rack-left, rack-right, right-top, and right-bottom.
There can only be one vial in each region at a time.
The position of the vials is determined by a motion capture system using structures placed on top of the vials as shown in Figure~\ref{fig:env_setup}F.
The state is the $x$ and $y$ location of all of the vials.



\subsection{Skills}

The robots are given a set of executable skills.

\textbf{Baxter Blocks:} 
The skills for the Baxter Blocks example are implemented as controllers that move the arm of the Baxter over the position of the block as determined by the AprilTags~\citep{Wang2016apriltags} attached to each block.
The skill then lifts up the block and moves it to over the destination location.
Finally, the skill lowers the block and releases it.

The left arm of the Baxter moves the red and blue blocks while the right arm moves the green block.
The skills allow the robot to move the red and blue blocks from D and E to A and AT (and vice-versa).
Another skill attempts to move the green block from F to C, but due to C being elevated, the skill sometimes results in the block ending in G.
Similarly, the skill from G to C sometimes results in the green block ending in C and sometimes in G.
Finally, there are skills that move the green block reliably from C to B and B to F.
Figure~\ref{fig:env_setup}A shows the skills available to the Baxter where skills with dashed lines have nondeterministic outcomes.

\begin{figure*}[t]
    \centering
    \includegraphics[width=\textwidth]{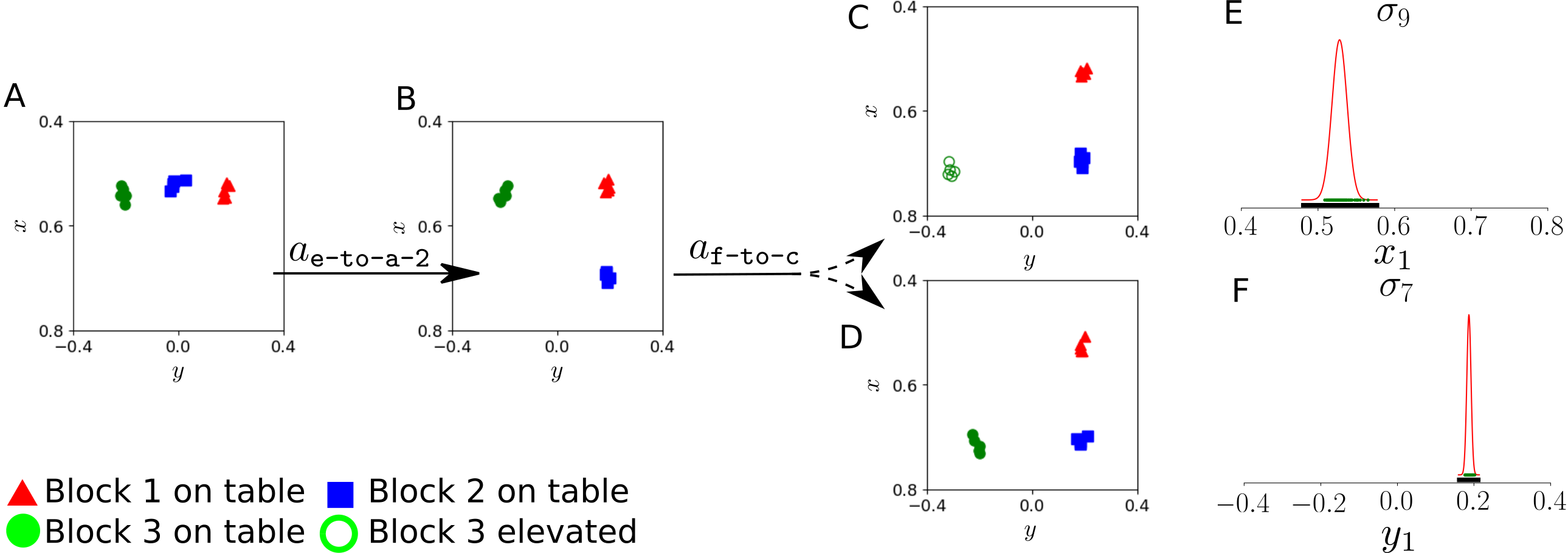}
    \caption{ Visualization of symbol combinations (A): $\sigma_{9} \wedge \sigma_{7} \wedge \sigma_{10} \wedge \sigma_{11} \wedge \sigma_{13} \wedge \sigma_{12} \wedge \sigma_{17} \wedge \sigma_{18} \wedge \sigma_{16}$, (B): $\sigma_{9} \wedge \sigma_{7} \wedge \sigma_{10} \wedge \sigma_{3} \wedge \sigma_{4} \wedge \sigma_{12} \wedge \sigma_{17} \wedge \sigma_{18} \wedge \sigma_{16}$, (C): $\sigma_{9} \wedge \sigma_{7} \wedge \sigma_{10} \wedge \sigma_{3} \wedge \sigma_{4} \wedge \sigma_{12} \wedge \sigma_{0} \wedge \sigma_{1} \wedge \sigma_{2}$, (D): $\sigma_{9} \wedge \sigma_{7} \wedge \sigma_{10} \wedge \sigma_{3} \wedge \sigma_{4} \wedge \sigma_{12} \wedge \sigma_{0} \wedge \sigma_{18} \wedge \sigma_{16}$. 
    All other symbols were $\false$. Ten samples were drawn from the intersection of the grounding sets of each symbol combination. Possible transitions are shown between the subfigures, corresponding to transitions in Equations \eqref{eq:e_to_a} and \eqref{eq:f_to_c}. Applying skill $a_\texttt{e-to-a-2}$ to (A) results in (B). 
    Applying skill $a_\texttt{f-to-c}$ in (B) results in (C) or (D). Examples of symbol groundings are shown in (E) and (F) in black circles. 
    The raw data is shown in green and the Gaussian fit to it in red.
    Figure adapted from \citet{pacheck2019automatic}.}
    \label{fig:robot_formula}
\end{figure*}

\textbf{Baxter Plates:} In the Baxter plates example, there are six skills that involve reproducing a trajectory demonstrated to the robot to move the green and blue plates from clean to set, set to dirty, and dirty to clean.
Skills require the plate to be in the named initial location in order to be executed (e.g. the blue plate must be clean in order for $a_\texttt{blue-clean-to-set}$ to be executed).
Some skills have additional restrictions on when they can be executed---the skill $a_\texttt{green-clean-to-set}$ cannot be executed when the blue plate is in set, $a_\texttt{green-dirty-to-clean}$ cannot be executed when the blue plate is dirty or clean, and $a_\texttt{blue-set-to-dirty}$ cannot be executed when the green plate is in set.
Figure~\ref{fig:env_setup}C shows the skills available to the Baxter and the approximate paths the plates follow between locations.

\textbf{Kinova Vials:} The skills in the Kinova Vials example allow the robot to move the vials between the yellow outer racks and the white rack (and vice-versa), but not between the two yellow racks.
For each skill, the arm moves between predetermined waypoints to move above the initial location, move down, grasp the vial, move up, move to above the destination location, move down, release the vial into the rack, move up, and then move back to a home position shown in Figure~\ref{fig:env_setup}F.
Figure~\ref{fig:env_setup}E shows the skills available to the Kinova arm; note that all the arrows are bi-directional.


\subsection{Symbol Generation}\label{sec:sym_gen}

We collected data to automatically generate symbols and encode the skills into an \gls{ltl} formula for each example. 

During the data collection process, an oracle tells the robot which skills can be executed based on the locations of the blocks, plates, or vials, and the robot randomly executes one of those skills.
In the Baxter Blocks and Plates examples, the oracle determines which skills can be executed based on the AprilTags~\citep{Wang2016apriltags} data, while the Kinova Vials example uses motion capture data.
Note that while the oracle in the Baxter Plates example uses AprilTags~\citep{Wang2016apriltags} to determine which skills can be executed, the symbol generation process uses only raw camera images.
For each example, we collected multiple precondition-skill-postcondition sets of data, along with which actions could be executed at each precondition.
Table~\ref{tab:demo_overview} lists the number of skill executions for each of the three examples.

\textbf{Baxter Blocks:} 
After collecting the data, we generated symbols and partitioned the skills for each of the examples \citep{konidaris2018skills}.
As shown in Table~\ref{tab:demo_overview}, we generated 19 symbols and 20 partitioned skills for the Baxter Blocks example.
The symbol generation process partitioned the skills that could move either the blue or red block from E or D to A or AT (and vice-versa) into two skills each.
The symbols generated refer to the $x$, $y$, or $z$ position for each of the blocks.

\textbf{Baxter Plates:} 
For the Baxter Plates example, we generated symbols directly from images taken by an external stationary USB camera, as shown in Figure~\ref{fig:env_setup}D.
To generate the symbols, we first resize the images to $120\times72$, convert them to grayscale and then apply independent component analysis \citep{hyvarinen2000independent}, keeping the top $5$ components.
The symbol generation process is then applied to these lower-dimensional vectors---preconditions are estimated using a support vector machine ($C=2, \gamma=4$) \citep{cortes1995support}, while effects are modelled using a kernel density estimator \citep{rosenblatt56,parzen1962estimation} with a Gaussian kernel and bandwidth determined by 3-fold cross validation. 
This procedure generates 5 factors and 9 symbols, where each symbol is a subset of the low-dimensional representation of an image.

\textbf{Kinova Vials:} 
In the Kinova Vials example, we generated 18 symbols and 48 partitioned skills.
Each symbol corresponds to a vial being located in a different region.
Due to the nature of the state space and skills, we were able to factor the state space such that each symbol is over both the $x$ and $y$ position of a vial, as opposed to only the $x$ or $y$ position, as in the Baxter Blocks example. 
Each skill in the Kinova Vials example changes both the $x$ and $y$ position of a vial; in the Baxter Blocks example the $x$, $y$, and $z$ position of blocks do not always change together.


\subsection{Skills-Based Specification}\label{sec:demo_robot_skills}

We automatically encode the symbols and skills in $\phifixed$ for each example.
We show selected parts from the specification for each example.

\textbf{Baxter Blocks:} In the Baxter Blocks example, we automatically encoded the symbols $\{\sigma_0,\ldots,\sigma_{18}\} \in \Sigma$ and skills $\{a_\texttt{f-to-c},\allowbreak\ldots,\allowbreak a_\texttt{d-to-at-2}\} \in \skills$ in $\phifixed$. 
In Equation \eqref{eq:c_to_p_pre}, we show part of the  system safety formula $\syssafetyfixed$.
We show part of the environment safety formula $\enveff$ in Equations \eqref{eq:e_to_a} and \eqref{eq:f_to_c}. 
Figure~\ref{fig:robot_formula}(A-D) visualizes the result of applying skills $a_\texttt{e-to-a-2}$ and $a_\texttt{f-to-c}$.

The precondition requirements of $a_\texttt{c-to-b}$ are encoded in $\syssafetyfixed$ in Equation~\eqref{eq:c_to_p_pre}.
\begin{equation}\label{eq:c_to_p_pre}
    \square(\lnot \bigcirc \sigma_1 \rightarrow \lnot \bigcirc a_\texttt{c-to-b})
\end{equation}

\begin{figure*}
\begin{equation}\label{eq:e_to_a}
    \begin{split}
        &\square (\ a_\texttt{e-to-a-2} \rightarrow
        (\bigcirc (\sigma_{3} \wedge \sigma_{4}) \wedge \bigcirc (\lnot\sigma_{11} \wedge \lnot\sigma_{13}) \bigwedge_{\sigma \in \effectsymstay{a_\texttt{e-to-a-2}}{1}} (\sigma \leftrightarrow \bigcirc \sigma)))
    \end{split}
\end{equation}
\begin{equation}\label{eq:f_to_c}
    \begin{split}
        &\square(a_\texttt{f-to-c} \rightarrow
        ((\bigcirc \sigma_0 \wedge \bigcirc \lnot\sigma_{17} 
        \bigwedge_{\sigma \in \effectsymstay{a_\texttt{f-to-c}}{1}} (\sigma \leftrightarrow \bigcirc \sigma)) 
        \vee \\ 
        & (\bigcirc (\sigma_0 \wedge \sigma_{1} \wedge \sigma_{2}) 
        \wedge \bigcirc (\lnot \sigma_{15} \wedge \lnot \sigma_{17} \wedge \lnot \sigma_{16} \wedge \lnot \sigma_{18})
        \bigwedge_{\sigma \in \effectsymstay{a_\texttt{f-to-c}}{2}} (\sigma \leftrightarrow \bigcirc \sigma))))
    \end{split}
\end{equation}
\end{figure*}
\begin{figure*}
\begin{equation}\label{eq:blue_clean_to_set_pre}
    \square (\lnot \bigcirc (\sigma_6 \wedge \sigma_3) \rightarrow \lnot \bigcirc a_\texttt{blue-clean-to-set})
\end{equation}
\begin{equation}\label{eq:blue_clean_to_set_post}
    \square (a_{\texttt{blue-clean-to-set}} \rightarrow (\bigcirc \sigma_5 \wedge \bigcirc \lnot \sigma_3 \bigwedge_{\sigma \in \effectsymstay{a_\texttt{blue-clean-to-set}}{1})} (\sigma \leftrightarrow \bigcirc \sigma)))
\end{equation}
\end{figure*}
\begin{figure*}
\begin{equation}\label{eq:green_rb_to_rl_pre}
\begin{split}
    &\square(\lnot \bigcirc ((\sigma_7 \wedge \sigma_0 \wedge \sigma_{13}) \vee (\sigma_7 \wedge \sigma_0 \wedge \sigma_{14}) \vee (\sigma_7 \wedge \sigma_0 \wedge \sigma_{15}) \vee (\sigma_7 \wedge \sigma_0 \wedge \sigma_{17}) \vee \\
    &(\sigma_{8} \wedge \sigma_0 \wedge \sigma_{13}) \vee (\sigma_8 \wedge \sigma_0 \wedge \sigma_{15}) \vee (\sigma_9 \wedge \sigma_0 \wedge \sigma_{13}) \vee (\sigma_9 \wedge \sigma_0 \wedge \sigma_{14}) \vee \\
    &(\sigma_9 \wedge \sigma_0 \wedge \sigma_{15}) \vee (\sigma_9 \wedge \sigma_0 \wedge \sigma_{17}) \vee (\sigma_{11} \wedge \sigma_0 \wedge \sigma_{14}) \vee (\sigma_{11} \wedge \sigma_0 \wedge \sigma_{15})) \rightarrow \\
    &\lnot \bigcirc a_\texttt{green-right-bottom-to-rack-left})
    \end{split}
\end{equation}
\begin{equation} \label{eq:green_rb_to_rl_post}
\begin{split}
    &\square(a_\texttt{green-right-bottom-to-rack-left} \rightarrow\\
    &(\bigcirc \sigma_4 \wedge \bigcirc (\lnot \sigma_0 \wedge \lnot \sigma_1 \wedge \lnot \sigma_2 \wedge \lnot \sigma_3 \wedge \lnot \sigma_5) \bigwedge_{\sigma \in \effectsymstay{a_\texttt{green-right-bottom-to-rack-left}}{1}}(\sigma \leftrightarrow \bigcirc \sigma))
    \end{split}
\end{equation}
\end{figure*}

Based on the data the robot has seen, it determines that it only needs to consider the value of $y_3$ in deciding if skill $a_\texttt{c-to-b}$ can be performed. 
There is only one symbol falling inside the precondition set so $\precondsym[a_\texttt{c-to-b}] = \{\{\sigma_1\}\}$. 
Therefore, equation \eqref{eq:c_to_p_pre} states that if $\sigma_1$ is not $\true$, i.e. block 3 is not at approximately $y=-0.3$m, skill $a_\texttt{c-to-b}$ can not be applied.

The part of $\envsafetyfixed$ pertaining to the effect of skill $a_\texttt{e-to-a-2}$ is shown in Equation~\eqref{eq:e_to_a} where $\effectsymstay{a_\texttt{e-to-a-2}}{1} = \{\sigma_0,\allowbreak \sigma_1,\allowbreak \sigma_2,\allowbreak \sigma_5,\allowbreak \ldots,\allowbreak \sigma_{10},\allowbreak \sigma_{12},\allowbreak \sigma_{14},\allowbreak \ldots, \sigma_{18}\}$. 
This corresponds to block 2 moving from location E to A and blocks 1 and 3 not moving. A potential outcome of applying skill $a_\texttt{e-to-a-2}$ is visualized in Figure \ref{fig:robot_formula}B.

The part of $\envsafetyfixed$ referring to the nondeterministic effects of skill $a_\texttt{f-to-c}$ is shown in Equation~\eqref{eq:f_to_c} where $\effectsymstay{a_\texttt{f-to-c}}{1} = \{\sigma_1, \ldots, \sigma_{16}, \sigma_{18}\}$ and $\effectsymstay{a_\texttt{f-to-c}}{2} = \{\sigma_3, \ldots, \sigma_{14}\}$. 
Equation \eqref{eq:f_to_c} encodes that when skill $a_\texttt{f-to-c}$ is applied, either $\sigma_0$ becomes $\true$ and $\sigma_{17}$ becomes $\false$ with symbols in $\effectsymstay{a_\texttt{f-to-c}}{1}$ not changing (block 3 ends in G), or $\sigma_{0}$, $\sigma_{1}$, and $\sigma_2$ become $\true$ and $\sigma_{15}$, $\sigma_{17}$, $\sigma_{16}$, and $\sigma_{18}$ become $\false$ with symbols in $\effectsymstay{a_\texttt{f-to-c}}{2}$ not changing (block 3 ends in C). 
This is visualized in Figures \ref{fig:robot_formula}C and D.

\begin{figure*}
    \centering
    \includegraphics[width=.9\textwidth]{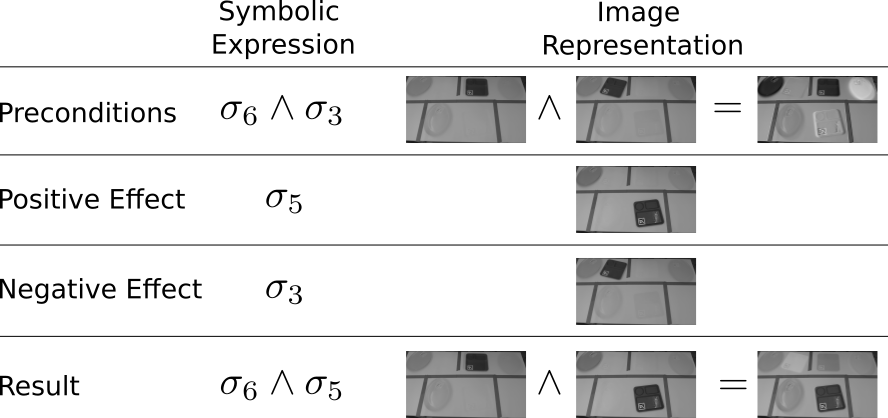}
    \caption{Symbolic representation of the precondition and effect for $a_{\texttt{blue-clean-to-set}}$.
The outcome of the action is computed by adding the positive effect to the precondition, and then removing the negative one. 
Since our representation is factorized, the preconditions and effects depend only on a subset of the factors that constitute the symbolic state space. 
In this case, $\sigma_3$ and $\sigma_5$ both refer to the same factor, while $\sigma_6$ refers to a different factor. The precondition for the skill represents states where the blue plate is in the clean position and the green plate is elsewhere. 
The effect of the skill is that the blue plate is now in the clean position; the green plate remains unaffected.
The location of the plates is entangled in the individual symbols, so we need to view combinations of symbols to know the location of the plates.}
    \label{fig:precondition_effect_plates}
\end{figure*}

\textbf{Baxter Plates:}
In the Baxter Plates example, the symbols correspond to images and the combination of symbols can be visualized together.
For example, Figure~\ref{fig:precondition_effect_plates} illustrates the preconditions and effects when the agent executes $a_{\texttt{blue-clean-to-set}}$.
The precondition requirements of $a_{\texttt{blue-clean-to-set}}$ are encoded in $\syssafetyfixed$ in Equation~\eqref{eq:blue_clean_to_set_pre}.
The postconditions of $a_{\texttt{blue-clean-to-set}}$ are encoded in $\envsafetyfixed$ as shown in Equation~\eqref{eq:blue_clean_to_set_post} where $\effectsymstay{a_\texttt{blue-clean-to-set}}{1} = \set{\sigma_0, \sigma_1, \sigma_2, \sigma_4, \sigma_6, \sigma_7, \sigma_8}$.

\textbf{Kinova Vials:}
For the Kinova Vials example, we also encode the preconditions and postconditions of the skills to move the vials in $\phifixed$.
We show part of the specification involving $a_\texttt{green-right-bottom-to-rack-left}$.
The preconditions of $a_\texttt{green-right-bottom-to-rack-left}$ are combinations of different locations of the red and yellow vials when the green vial is in right-bottom and the rack-left location is clear as shown in Equation~\eqref{eq:green_rb_to_rl_pre}.
For all of the Kinova Vials skills, the skills could have up to 16 preconditions, as the precondition classifier does not necessarily learn that the vials not moving cannot be in the same physical location.
In this example, there are only 12 preconditions, with one $(\sigma_7 \wedge \sigma_0 \wedge \sigma_{13})$ having both the red and yellow vials at the right-top location.

The postconditions of $a_\texttt{green-right-bottom-to-rack-left}$ make the green vial in the rack-left location and not in the right-bottom location without changing the position of the red and yellow vials as shown in Equation~\eqref{eq:green_rb_to_rl_post} where $\effectsymstay{a_\texttt{green-right-bottom-to-rack-left}}{1} = \{\sigma_6, \sigma_7, \ldots, \sigma_{16}, \sigma_{17}\}$.


\subsection{Realizable Base Task Specifications}\label{sec:demo_task}

\begin{figure*}
\centering
\includegraphics[width=0.7\textwidth]{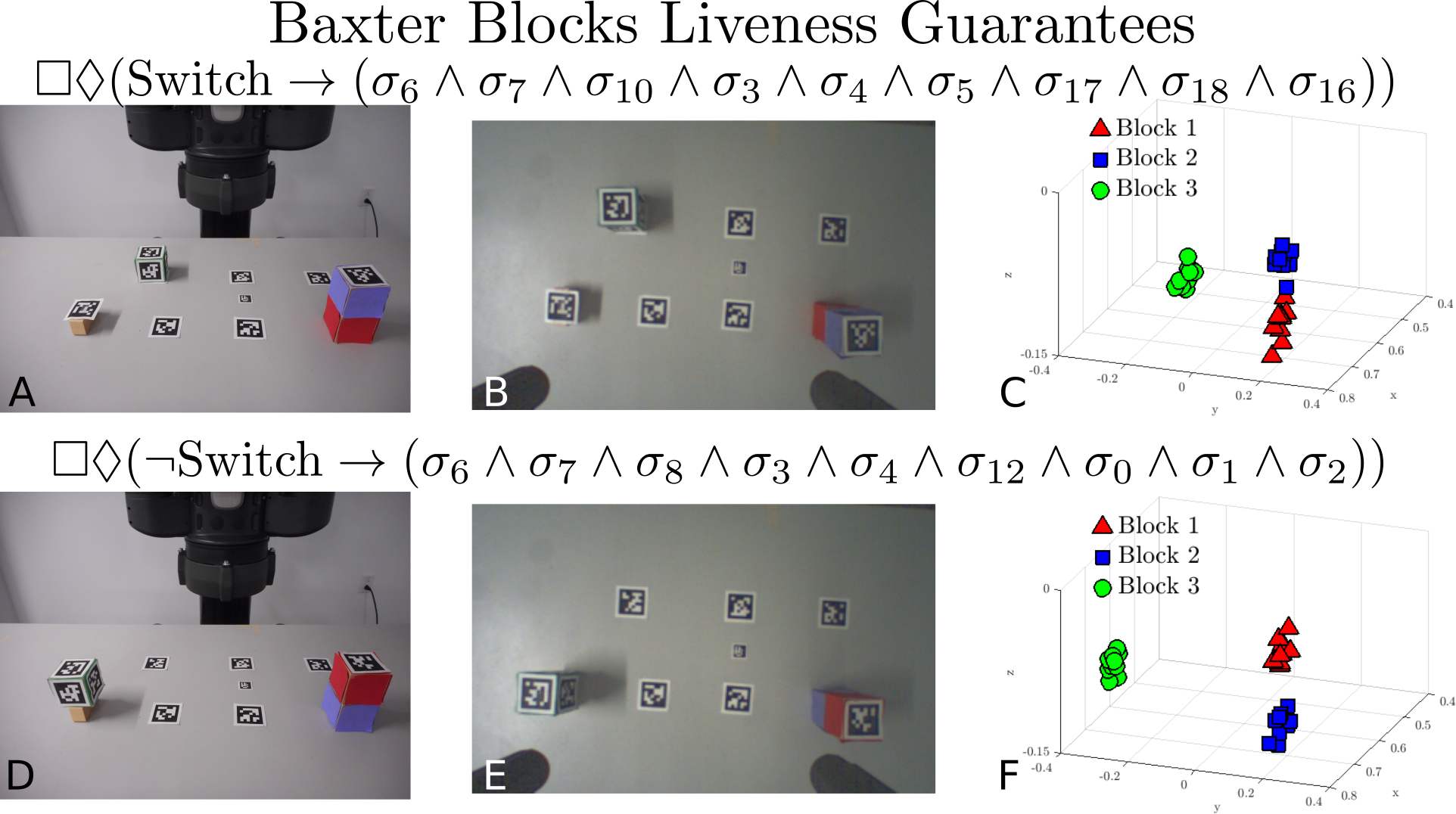}
\caption{
In the Baxter Blocks example, the robot reacts to the value of $\switch \in \user$.
When $\switch=\true$, the robot needs to stack the blue block on the red block in location A and the green block in location F. 
When $\switch=\false$, the robot needs to stack the red block on the blue block in location A and the green block in location C. 
Figure from \citet{pacheck2019automatic}.
}
\label{fig:blocks_liveness}
\end{figure*}

\begin{figure*}
\centering
\includegraphics[width=\textwidth]{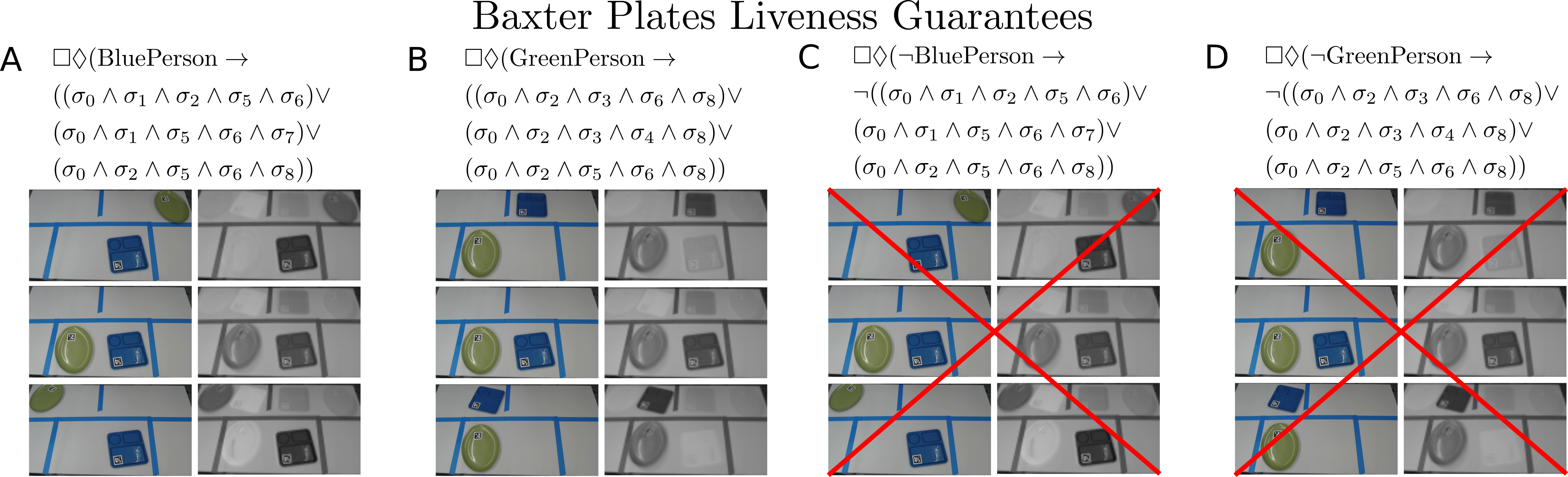}
\caption{
In the Baxter Plates example, 
A) when $\blueperson=\true$ the blue plate should be set, 
B) when $\greenperson=\true$ the green plate should be set, 
C) when $\blueperson=\false$ the blue plate should not be set, and 
D) when $\greenperson=\false$ the green plate should not be set.
The left column of each subfigure shows the physical interpretation of the liveness guarantee and the right column shows the corresponding image generated based on the symbols.}
\label{fig:plates_liveness}
\end{figure*}

\begin{figure*}
\centering
\includegraphics[width=\textwidth]{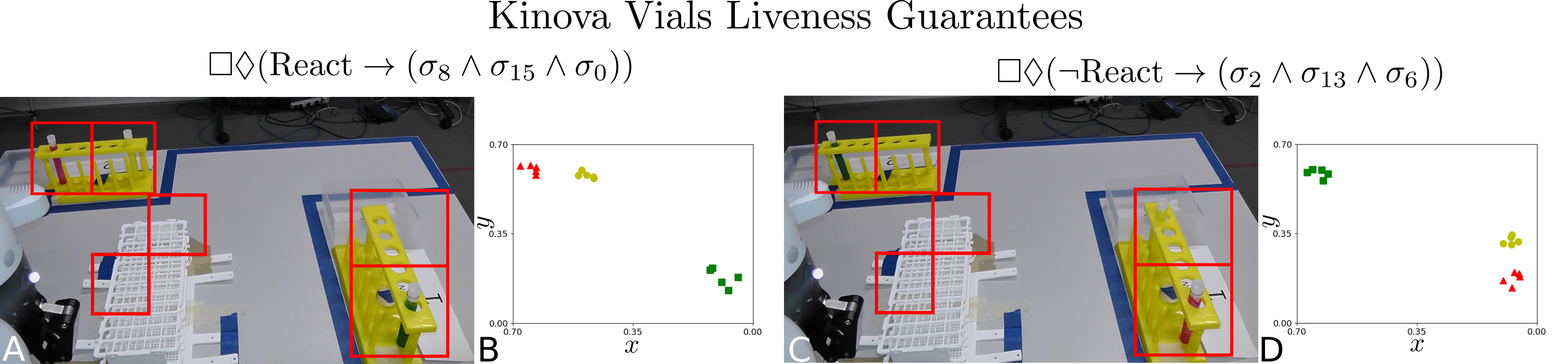}
\caption{
In the Kinova Vials example, the robot reacts the the truth value of $\react \in \user$.
When $\react=\true$, the red vial should be in the top-left location, the yellow vial in the top-right location, and the green vial in the right-bottom location.
When $\react=\false$, the red vial should be in the right-bottom location, the yellow vial in the right-top location, and the green vial in the top-left location.}
\label{fig:vials_liveness}
\end{figure*}

We introduce additional environment variables $\varuser \in \user$ that the user controls.
Using $\user$, we write reactive tasks.
We describe the user-defined task for each example and show parts of the specification in this section.
We show the full $\phifixed$ for each example in Appendix~\ref{sec:appendix}.

\textbf{Baxter Blocks:} 
For the Baxter Blocks example, we introduce an additional environment variable $\user=\{\switch\}$. 
The task liveness specifications are shown in Figure~\ref{fig:blocks_liveness}.
When $\switch=\true$, the red block (block 1) should eventually be in A, the blue block (block 2) in AT, and the green block (block 3) in F as shown in Figure~\ref{fig:blocks_liveness}(A-C) and encoded in the \gls{ltl} formula above the subfigures. 
When $\switch=\false$, the red block (block 1) should be in AT, the blue block (block 2) in A, and the green block (block 3) in C as shown in Figure~\ref{fig:blocks_liveness}(D-F) and encoded in the \gls{ltl} formula above the subfigures.

We include a fairness assumption on the environment that the green block (block 3) will eventually be placed in location C when $a_\texttt{g-to-c}$ is applied shown in Equation~\eqref{eq:env_live}. 
Without this, the specification is unrealizable because in the worst case, skill $a_\texttt{g-to-c}$ always results in the green block (block 3) ending in G.
\begin{equation}\label{eq:env_live}
    \envlivetask = \square \lozenge (a_\texttt{g-to-c} \rightarrow (\sigma_0 \wedge \sigma_1 \wedge \sigma_2))
\end{equation}

\textbf{Baxter Plates:} 
For the Baxter Plates example, we introduce two additional user-defined variables $\user=\{\blueperson,\ \greenperson\}$.
The task liveness specification is to make the blue plate set when $\blueperson = \true$, the green plate set when $\greenperson = \true$, the blue plate not set when $\blueperson = \false$, and the green plate not set when $\greenperson = \false$ as shown and encoded in Figure~\ref{fig:plates_liveness}A, B, C, and D respectively.
Note that in the Baxter Plates example, we need to specify the location of both plates in the liveness guarantees due to the entanglement of the symbols.

The symbol generation process was not able to fully determine the effects of two of the skills, due to the lossy nature of the compressed state representation.
The symbol generation process learns that for the skills $a_\texttt{green-clean-to-set}$ and $a_\texttt{blue-clean-to-set}$, a possible outcome is that no plates move.
We add a fairness assumption that the skills should always eventually succeed in moving the plates shown in Equation~\eqref{eq:plates_env_liveness}.

\begin{equation}\label{eq:plates_env_liveness}
    \begin{split}
        &\square \lozenge (a_\texttt{green-clean-to-set} \rightarrow \sigma_8) \wedge \\
        &\square \lozenge (a_\texttt{blue-clean-to-set} \rightarrow \sigma_5)
    \end{split}
\end{equation}

\textbf{Kinova Vials:} For the Kinova Vials example, we introduce the additional environment variable $\user=\{\react\}$.
The task is to arrange the vials in one configuration when $\react=\true$ and another when $\react=\false$.
When $\react = \true$, the Kinova should arrange the vials such that the green vial is in the right-bottom position, the red vial is in the top-left position, and the yellow vial is in the top-right position (Figure~\ref{fig:vials_liveness}(A,B)).
When $\react = \false$, the Kinova should arrange the vials such that the green vial is in the top-left position, the red vial is in the right-bottom position, and the yellow vial is in the right-top position.
The \gls{ltl} formula encoding the task liveness specification is shown above the visual interpretation of the task in Figure~\ref{fig:vials_liveness}.

We add an additional constraint to $\syssafetytask$ that the red and green vials should never be in the same yellow rack as shown in Equation~\eqref{eq:vials_safety_task} and Figure~\ref{fig:vaccine_fault_model}(A,B,D,E).
\begin{equation}\label{eq:vials_safety_task}
\begin{split}
    \syssafetytask = &\square \lnot (\sigma_0 \wedge \sigma_7) \wedge \square \lnot (\sigma_1 \wedge \sigma_6) \wedge \\
    & \square \lnot (\sigma_2 \wedge \sigma_9) \wedge \square \lnot (\sigma_3 \wedge \sigma_8) \wedge \\
    &\square \lnot \bigcirc (\sigma_0 \wedge \sigma_7) \wedge \square \lnot \bigcirc (\sigma_1 \wedge \sigma_6) \wedge \\
    & \square \lnot \bigcirc (\sigma_2 \wedge \sigma_9) \wedge \square \lnot \bigcirc (\sigma_3 \wedge \sigma_8)
    \end{split}
\end{equation}


\subsection{Synthesis and Execution}

\begin{figure*}[t]
\centering
\includegraphics[width=\textwidth]{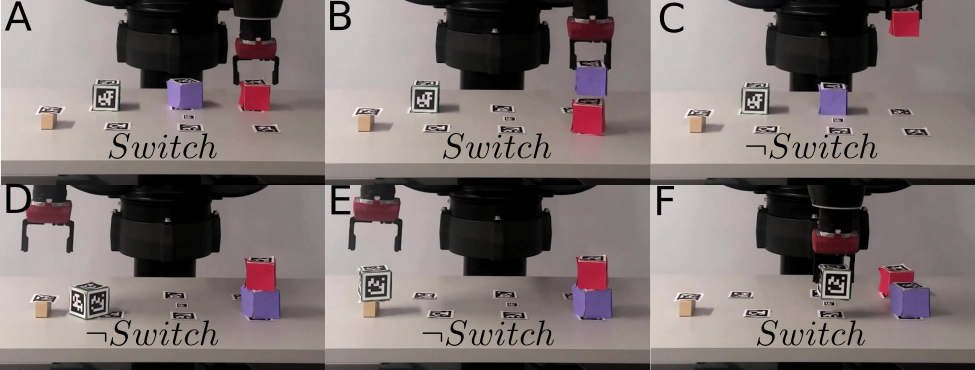}
\caption{
The Baxter robot executing a strategy to fulfill the desired liveness guarantees in Figure~\ref{fig:blocks_liveness} without any additional user provided constraints.
The value of $\switch$ was controlled through a user interface. Figure adapted from \citet{pacheck2019automatic}.
}
\label{fig:sequence}
\end{figure*}

For each example, we are able to find a strategy to fulfill the original task.
Throughout this paper, computation times refer to running Slugs \citep{ehlers2016slugs} and our algorithms on an Ubuntu 18.04 machine with 12 GB RAM.
For the Baxter Blocks example, we synthesize $\strat$ with 256 states in 1 second. 
For the Baxter Plates example, it took 1 second to synthesize $\strat$ with 124 states.
The strategy for the Kinova Vials example took 22 seconds to synthesize and had 232 states.

\begin{figure*}
    \centering
    \includegraphics[width=.8\textwidth]{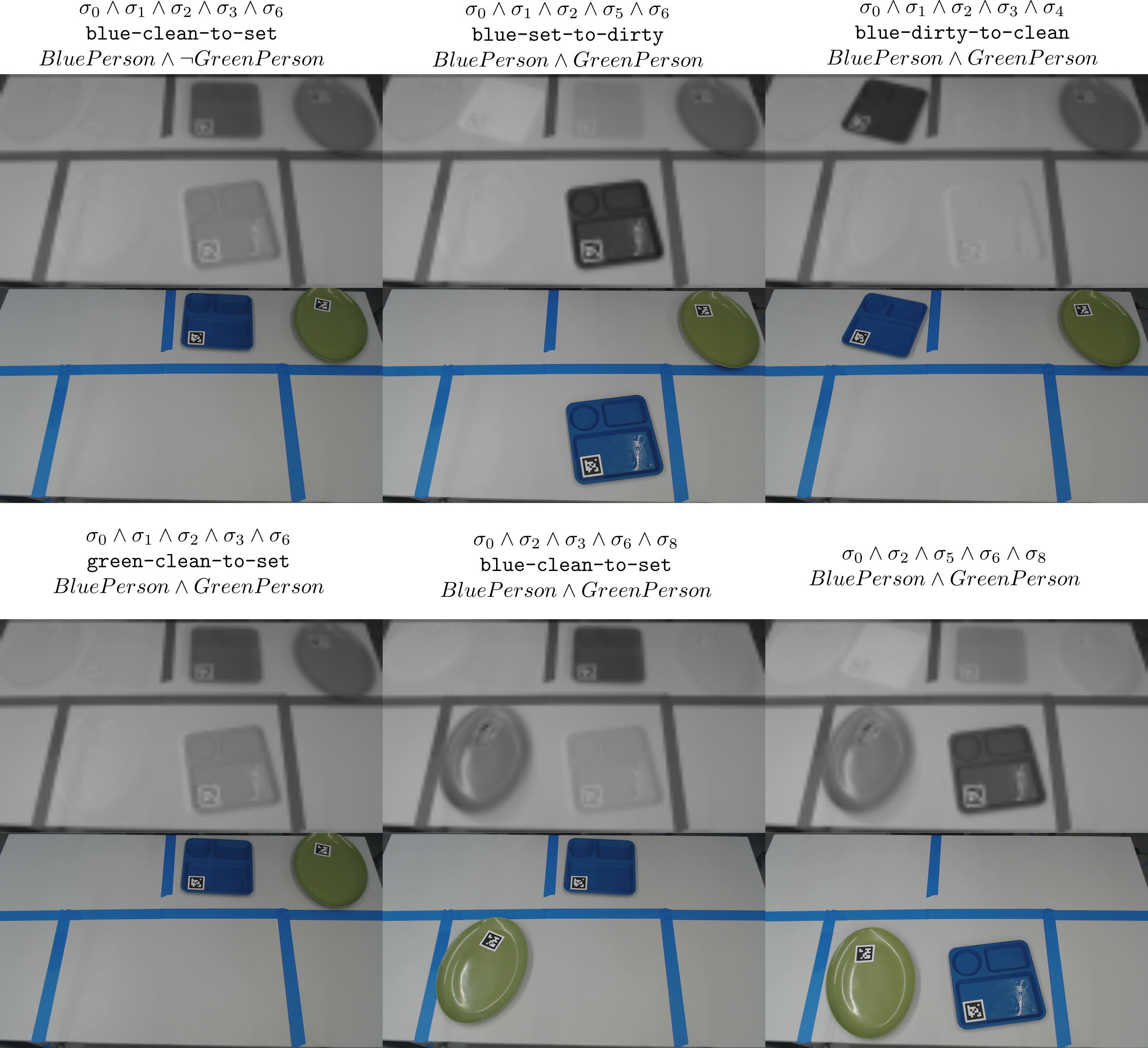}
    \caption{
    Sequence of states needed to achieve the liveness guarantees in Figure~\ref{fig:plates_liveness} for the Baxter Plates example.
    The strategy gives skills that react to the truth value of $\blueperson$ and $\greenperson$.
    We show the physical and symbolic interpretation of each state. 
    The upper image of each pair is the visualization of the symbolic state, created by combining the symbols that are $\true$. 
    The lower image of each pair is the physical state.
    Above each pair of images, we show which symbols are $\true$, which action should be taken, and the truth value of $\blueperson$ and $\greenperson$.}
    \label{fig:plates_sequence}
\end{figure*}

We demonstrate the strategy for the Baxter Blocks example.
We controlled the value of $\switch \in \user$ through a user interface.
We sampled the current state $x \in X$ to find out which symbols were $\true$. 
A symbol $\sigma_{a,j,f_q}$ was $\true$ if the state was in the grounding set for the symbol, $\mathcal{G}(\sigma_{a,j,f_q})$.
All other symbols were $\false$. 
We show an example execution of $\strat$ for the Baxter Blocks example in Figure~\ref{fig:sequence}.
We show an example sequence of states in $\strat$ for the Baxter Plates example in Figure~\ref{fig:plates_sequence}.

\subsection{Repair of Unrealizable Tasks}

We demonstrate the repair process by finding skill suggestions for six unrealizable specifications.
For each example, we made the specification unrealizable by either adding additional task constraints to $\syssafetytask$ or modifying $\syssafetyfixed$.
Each unrealizable specification shows different aspects of the repair process.

For the Baxter Blocks example, we investigate four unrealizable specifications.
In two specifications, we add constraints to avoid a skill or set of states.
In these two specifications we find $\skills_\textrm{n.o.t} \neq \varnothing$, allowing us to narrow the search space for new skills to those with the same preconditions as $a \in \skills_\textrm{n.o.t}$. 
For two other specifications, we modify the skills available to the robot and find $\skills_\textrm{n.o.t} = \varnothing$, requiring us to perform an exhaustive search for new skills over all current preconditions sets.

For the Baxter Plates and Kinova Vials examples, we investigate one unrealizable specification each.
In the Baxter Plates example, we add a reactive task constraint and show the benefits of the enumeration-based repair approach over the synthesis-based approach.
In the Kinova Vials example, we add a constraint to avoid certain states and show the benefits of the synthesis-based repair approach over the enumeration-based approach.
For these unrealizable specifications, we find $\skills_\textrm{n.o.t} \neq \varnothing$.
The Baxter Plates and Kinova Vials examples both required two skills to repair, which increased the complexity of the repair process.

We describe the constraints added to each example that make the specifications unrealizable in this section and show selected formula.
We show the full unrealizable task specifications, $\phifixed$, in Appendix~\ref{sec:appendix}.

We give an overview of the number of solutions and time taken to find those solutions in Table~\ref{tab:repair_overview}.

\begin{table*}[]
\centering
\begin{tabular}{@{}cccccc@{}}
\toprule
                               &                   & \multicolumn{2}{c}{Enumeration-based}                 & \multicolumn{2}{c}{Synthesis-based}                   \\ \midrule
                               & \makecell{Unrealizable\\spec} & \makecell{Number of\\suggestions}     & Time (sec)    & \makecell{Number of\\suggestions}     & Time (sec)                \\ \midrule
\multirow{4}{*}[-7pt]{Baxter Blocks}  & 1                 & 6 & 163.2 & 50 & 70 \\ \cmidrule(l){2-6} 
                               		  & 2                 & 3 & 171.6 & 36 & 47 \\ \cmidrule(l){2-6} 
                               	  	  & 3                 & 68 & 6420 & 32 & 49 \\ \cmidrule(l){2-6} 
                               		  & 4                 & 17 & 2868 & 35 & 32 \\ \midrule
Baxter Plates                  		  & 5                 & 9 & 1214 & 1 & 1.1 \\ \midrule 
Kinova Vials                   		  & 6                 & 22 & 25167 & 25 & 381 \\ \bottomrule
\end{tabular}
\caption{Overview of the repair suggestions found. The synthesis-based repair approach takes substantially less time than the enumeration-based repair approach. For the Kinova Vials example, we were not able to run the enumeration-based repair approach until completion; the numbers listed are for stopping the repair process midway.}
\label{tab:repair_overview}.
\end{table*}


\begin{figure*}
    \centering
    \includegraphics[width=\textwidth]{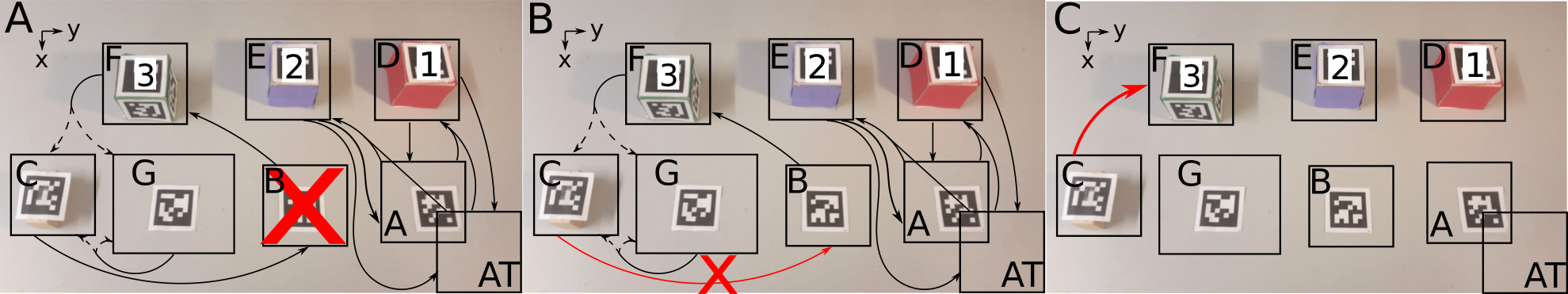}
    \caption{(A) The added constraint in Unrealizable Specification 1 that the green block (block 3) should never be in location B.
    (B) The added constraint in Unrealizable Specification 2 that the skill \texttt{c-to-b} should never be executed. 
    (C) A skill moving the green block (block 3) from location C to location F makes both Unrealizable Specification 1 and 2 (and 3 and 4) realizable. 
    Figure from \citet{pacheck2019automatic}.}
    \label{fig:block_fault_model}
\end{figure*}

\begin{figure*}
    \centering
    \includegraphics[width=\textwidth]{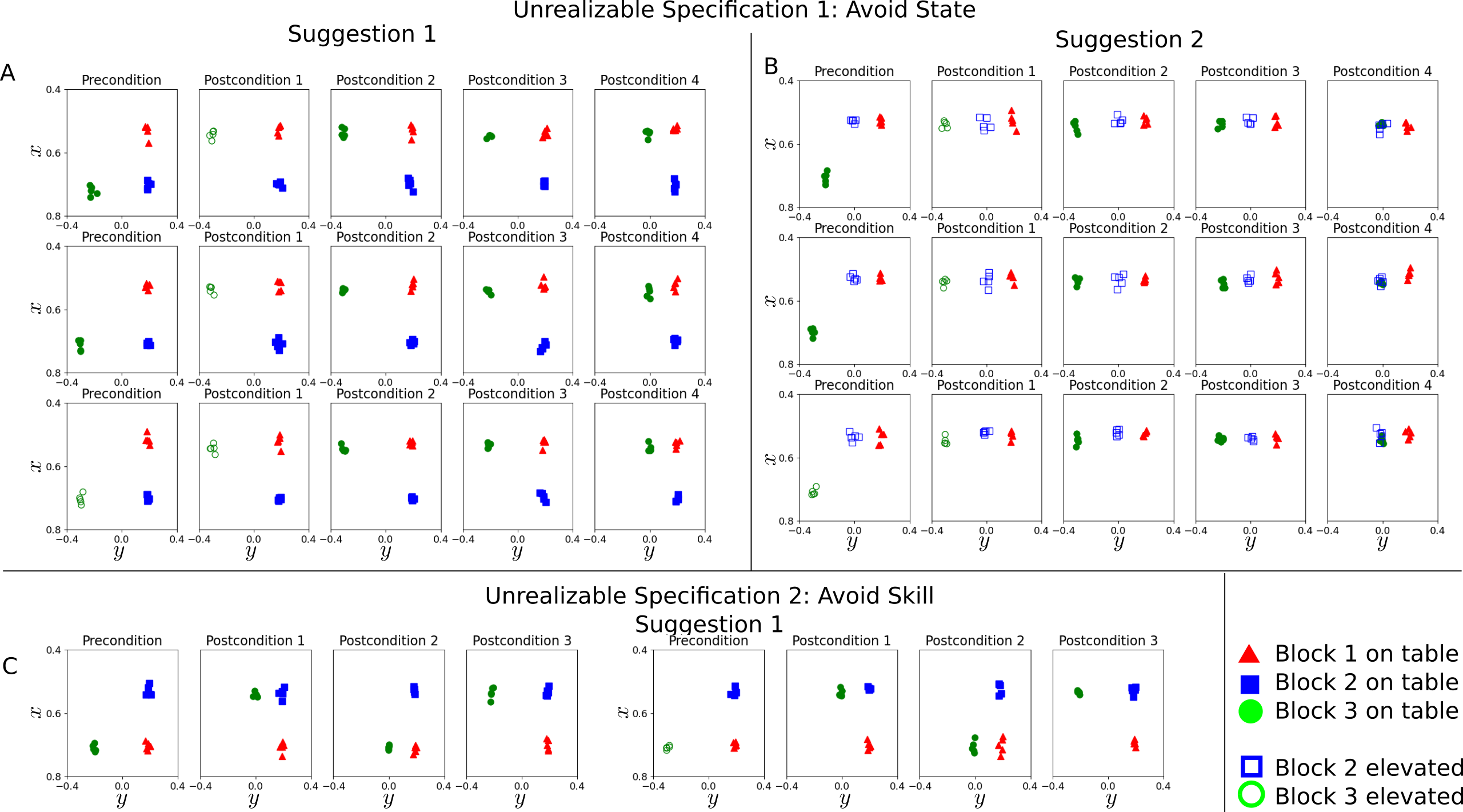}
    \caption{Selected suggestions from the synthesis-based repair process for Unrealizable Specifications 1 and 2.
    A) Suggestion 1 for Unrealizable Specification 1 proposes three skills with nondeterministic postconditions.
    B) Suggestion 2 for Unrealizable Specification 1 proposes three skills.
    These skills have the precondition and postcondition that block 2 is hovering over location E.
    This is physically impossible, but the specification does not disallow it.
    C) Suggestion 1 for Unrealizable Specification 2 proposes two skills with nondeterministic postconditions.}
    \label{fig:synthesis_suggestions_unrealizable_1_and_2}
\end{figure*}

\textbf{Baxter Blocks:} The two Baxter Blocks unrealizable specifications for which we can narrow the search space of possible skills have the same $\phifixed$ and $\phitask$ as in Section \ref{sec:demo_robot_skills} and \ref{sec:demo_task}, with the addition of $\syssafetytask$, as shown in Figure~\ref{fig:block_fault_model} and described below.
For each specification, we add the constraint that only one block can move at a time as shown in Appendix~\ref{sec:appendix_baxter_blocks}.

\textbf{Unrealizable Specification 1}: 
In Figure~\ref{fig:block_fault_model}A, we show the added constraint that the green block (block 3) never be in location B, $\syssafetytask = \square \lnot (\sigma_{0} \wedge \sigma_{15}) \wedge \square \lnot \bigcirc (\sigma_{0} \wedge \sigma_{15})$. 
This type of scenario could occur if there was an obstacle in location B.

\textbf{Unrealizable Specification 2}: 
In Figure \ref{fig:block_fault_model}B, we show the added constraint that the robot never use skill $a_\texttt{c-to-b}$, $\syssafetytask = \square \lnot a_\texttt{c-to-b} \wedge \square \lnot \bigcirc a_\texttt{c-to-b}$. 
This type of scenario could occur if a motor enabling skill $a_{\texttt{c-to-b}}$ was damaged and the skill could not be performed.

\textbf{Unrealizable Specifications 1 and 2 Repair}: 
For both Unrealizable Specifications 1 and 2, the enumeration-based repair process found $\skills_\textrm{n.o.t} = \{a_\texttt{c-to-b}\}$, corresponding to the precondition that the green block (block 3) be in location C. 
We only searched for one additional skill, so we did not need to find combinations of skills in Line~\ref{eq:combinations_alg} of Algorithm~\ref{alg:repair1}.
The repair process searched through 62 skills to find six skill suggestions for Unrealizable Specification 1 and three skill suggestions for Unrealizable Specification 2 in 2.7 minutes and 2.9 minutes, respectively. 
The user needs to determine which suggestion to implement.
Some suggestions were not physically possible, making them impossible to implement, such as a suggestion with $\precondsym[\anew] = \{\{\sigma_1\}\}$ and $\effectsym{\anew}{} = \{\sigma_{17},\sigma_{1},\sigma_{16}\}$, corresponding to moving the green block (block 3) to the $x$ position of location F and the $y$ and $z$ position of location C, which would leave the block floating in the air.
If desired, the user could add additional constraints to the specification to remove the physically impossible suggestions.
One skill suggestion for both Unrealizable Specifications 1 and 2, with $\precondsym[\anew] = \{\{\sigma_1\}\}$ and $\effectsym{\anew}{} = \{\sigma_{17},\sigma_{18},\sigma_{16}\}$, corresponding to moving the green block (block 3) from location C to F, is physically possible. 
When this skill is added to Unrealizable Specifications 1 and 2, the task is realizable. 

The synthesis-based repair process found 50 suggestions in 70 seconds for Unrealizable Specification 1 and 36 suggestions in 47 seconds for Unrealizable Specification 2.
Figure~\ref{fig:synthesis_suggestions_unrealizable_1_and_2} shows selected suggestions to repair Unrealizable Specifications 1 and 2.
The repair suggestions for Unrealizable Specifications 1 and 2 involve adding one or more new skills or relaxing the preconditions.

The synthesis-based repair process produces suggestions that exploit all possible preconditions of the existing skills.
Suggestion 1 proposes 3 skills to repair Unrealizable Specification 1.
For all of the skills in Suggestion 1, the red block (block 1) is in location D and blue block (block 2) is in location A.
Each row in Figure~\ref{fig:synthesis_suggestions_unrealizable_1_and_2}A shows one skill.
The first skill moves the green block (block 3) from location G to either floating above a new location with the $x$ position of F and the $y$ position of C (Postcondition 1), a new location with the $x$ position of F and the $y$ position of C (Postcondition 2), F (Postcondition 3), or D (Postcondition 4).
The other two skills move the green block (block 3) from on the table in location C or stacked in location C.

Only Postcondition 3 was seen during the original symbol learning. 
However, the synthesis-based repair process exploits the existing preconditions and postconditions of other skills when it creates or modifies skills, enabling the robot to create behaviors not observed before.

The remainder of the suggestions to repair Unrealizable Specification 1 involving adding new skills  are similar to Suggestion 1, with multiple skills that move the green block (block 3) to different locations that may not be physically possible, but still fall in the precondition of skill $a_\texttt{f-to-c}$.
The other suggestions have different configurations of the red and blue blocks (blocks 1 and 2).
Since we do not add constraints that different blocks can not be in the same physical position or floating in the air, some suggestions may be physically impossible.
Suggestion 2 for Unrealizable Specification 1 in Figure~\ref{fig:synthesis_suggestions_unrealizable_1_and_2}B shows a suggestion with the blue block (block 2) in location E, but with a $z$ value that is above the table.

The synthesis-based repair process generates similar suggestions for Unrealizable Specification 2.
Suggestion 1 proposes two skills to make the specification realizable.
One skill moves the green block (block 3) from location G to either location B, E, or F.
The other skill moves the green block (block 3) from location C to to either location B, E, or F.

The suggestions that relax the preconditions of the skills for Unrealizable Specification 1 and 2 are physically impossible.
These involve adding preconditions that allow for blocks to be floating in the air or at the same location as other blocks, which is not possible.
We show examples of preconditions being relaxed in the Kinova Vials Unrealizable Specification 6.
If desired, additional constraints could be added to the specification to generate suggestions without these physically impossible suggestions.


Unrealizable Specifications 3 and 4 involved modifying the base specification.

\begin{figure*}
    \centering
    \includegraphics[width=\textwidth]{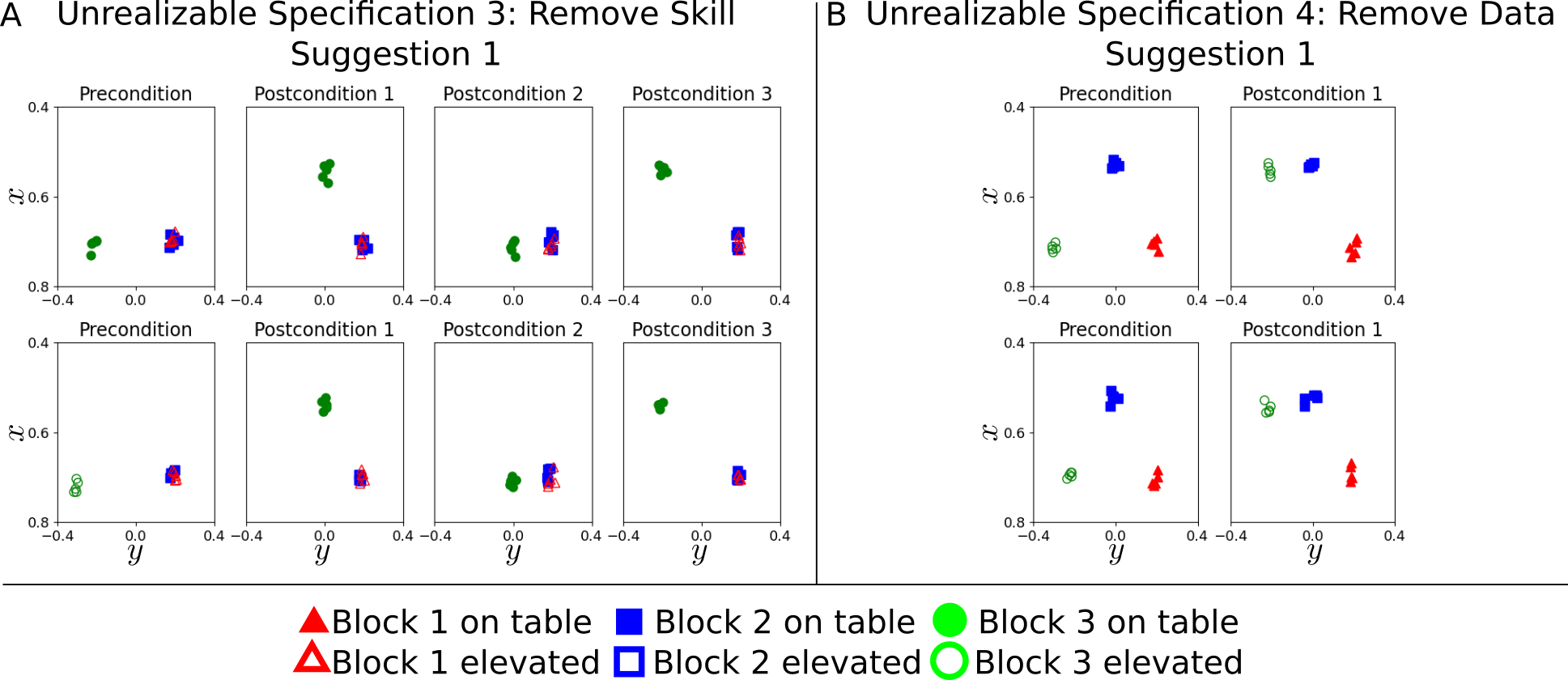}
    \caption{Selected suggestions from the synthesis-based repair process for Unrealizable Specifications 3 and 4.
    (A) Two additional skills to move block 3 that have nondeterministic postconditions of locations B, E, and F. One skill starts from location C and one from location G. 
    (B) One suggestion to repair Unrealizable Specification 4 involves two new skills. One moves block 3 from location G to F and the other moves block 3 from location C to F.}
    \label{fig:synthesis_suggestions_unrealizable_3_and_4}
\end{figure*}

\textbf{Unrealizable Specification 3}: 
We removed skill $a_\texttt{c-to-b}$ from $\skills$ before writing the specification, using the same set of symbols $\Sigma$ as in Section \ref{sec:sym_gen}. 
The user defined task was the same as in Figure~\ref{fig:blocks_liveness}.

\textbf{Unrealizable Specification 4}: 
We removed all data pertaining to skill $a_\texttt{c-to-b}$ before the symbol generation process. 
This resulted in a different set of symbols, $\Sigma$. 
The user defined task was the same as represented in Figure \ref{fig:blocks_liveness}. 
There were no longer symbols corresponding to the green block (block 3) being in location B, as symbols are only generated from effect sets, so the subscripts of the symbols in the liveness guarantees shown in Figure~\ref{fig:blocks_liveness} were different.

\textbf{Unrealizable Specification 3 and 4 Repair}: For both Unrealizable Specification 3 and 4, the enumeration-based repair process found $\skills_\textrm{n.o.t} = \varnothing$, requiring an exhaustive search of the skill space. 
For Unrealizable Specification 3, the repair process searched through 1172 skills and found 68 possible new skills in 107 minutes.
For Unrealizable Specification 4, the repair process searched through 788 skills and found 17 possible skills to repair the specification in 48 minutes. 
The repair process suggested a skill that would move the green block (block 3) from both locations C and G to location F for both specifications. 
With the fairness assumption in Equation~\eqref{eq:env_live}, this has the same result as giving the robot a skill moving the green block (block 3) from C to F. 
When we added a skill that moved the green block (block 3) from C to F, both specifications were realizable.

The synthesis-based repair found 32 suggestions to repair Unrealizable Specification 3 in 49 seconds and 35 suggestions to repair Unrealizable Specification 4 in 32 seconds.
Figure~\ref{fig:synthesis_suggestions_unrealizable_3_and_4} shows selected suggestions from the synthesis-based repair process to repair Unrealizable Specifications 3 and 4.

We show one suggestion to repair Unrealizable Specification 3 in Figure~\ref{fig:synthesis_suggestions_unrealizable_3_and_4}A.
This suggestion is similar to those for Unrealizable Specifications 1 and 2.
The repair process suggests two new skills that move the green block (block 3), one that has a precondition of location C and one that has a precondition of location G.
Both skills have postconditions that move to either location B, E, or F.
The remainder of the suggestions to repair Unrealizable Specification 3 are similar but have different configurations of the red and blue blocks (blocks 1 and 2).
Again, the suggestions involving relaxing the preconditions are physically impossible.

One of the suggestions to repair Unrealizable Specification 4 is shown in Figure~\ref{fig:synthesis_suggestions_unrealizable_3_and_4}B.
This suggestion proposed two skills to move the green block (block 3) to location F from either C or G.
Note that since we removed data to generate Unrealizable Specification 4, there is only a symbol pertaining to the green block (block 3) being elevated off the table.


\begin{figure*}
    \centering
    \includegraphics[width=\textwidth]{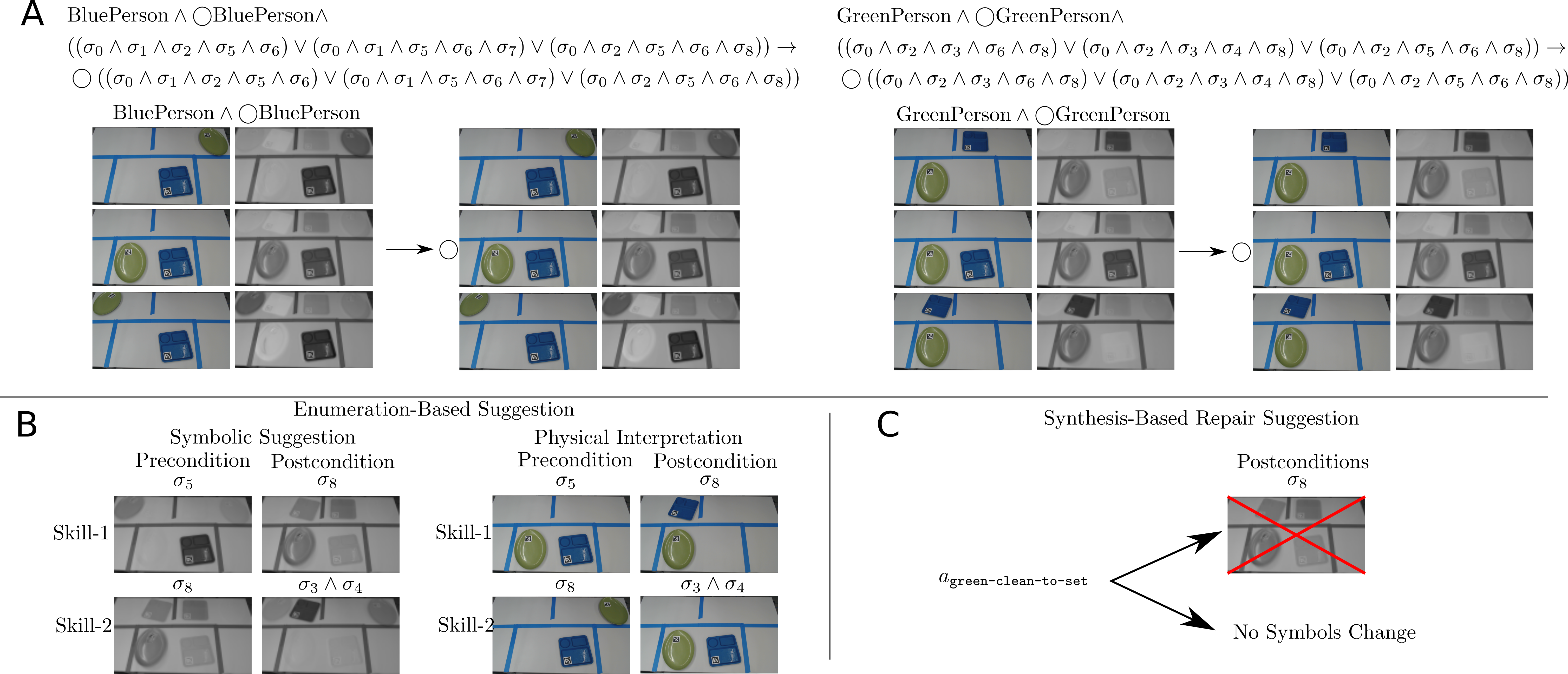}
    \caption{(A) The added constraint in Unrealizable Specification 5 for the Baxter Plates example.
    When $\blueperson$ is $\true$ at both the current and next step and the blue plate is set, the blue plate should be set at the next step.
    Similarly, when $\greenperson$ is $\true$ at both the current and next step and the green plate is set, the green plate should be set at the next step.
    (B) One of the 9 suggestions found by the enumeration-based repair process.
    The suggestion proposes two new skills.
    The first skill moves the blue plate from set to dirty when the green plate is set.
    The second skill moves the green plate from clean to set when the blue plate is set.
    (C) The synthesis-based repair process finds one suggestion to repair the specification.
    This suggestion proposes modifying the skill $a_\texttt{green-clean-to-set}$ such that it does nothing to violate the added environment liveness assumption.}
    \label{fig:plate_fault_model}
\end{figure*}

\textbf{Baxter Plates:} 
For the Baxter Plates example, we investigate one unrealizable specification.

\textbf{Unrealizable Specification 5}: 

The user-defined liveness guarantees are the same as in Figure~\ref{fig:plates_liveness}.
We add the additional constraints that the when either $\blueperson$ or $\greenperson$ is $\true$, the same colored plate should not be moved out of the set position.
This is encoded in $\syssafetytask$ as shown in Figure~\ref{fig:plate_fault_model}A.

\textbf{Unrealizable Specification 5 Repair}:
The enumeration-based repair process found $\skills_\textrm{n.o.t} \neq \emptyset$.
We considered repair suggestions that consisted of two additional skills.
The repair process searched through 3840 potential new skills and found 9 suggestions in 20.23 minutes.
We show one of the suggestions in Figure~\ref{fig:plate_fault_model}B.
The first skill has the precondition $\sigma_5$ and postcondition $\sigma_8$.
When this skill is executed in the repaired strategy, this corresponds to the blue plate moving from set to dirty when the green plate is set.
The second skill has the precondition $\sigma_8$ and postconditions $\sigma_3 \wedge \sigma_4$.
This corresponds to moving the green plate from clean to set when the blue plate is set.

The synthesis-based repair process found 1 suggestion in 1.1 seconds.
This suggestion is to reduce the nondeterminism in skill $a_\texttt{green-clean-to-set}$ such that it does nothing.
This suggestion is valid symbolically because it works to violate the liveness assumption that the skill $\square \lozenge (a_\texttt{green-clean-to-set} \rightarrow \sigma_8)$.

We investigated if the synthesis-based repair would find suggestions for repair if the symbol generation process was able to determine the skill symbolic structure correctly.
We modified the learned skills to remove the nondeterminism, removing the option for skills $a_\texttt{green-clean-to-set}$ and $a_\texttt{blue-clean-to-set}$ to not change any environment variables.

The synthesis-based repair was unable to find any other suggestions for repair.
During the synthesis-based repair process, Algorithm~\ref{alg:restrictPost}, \texttt{restrictPostconditions}, finds the required postconditions for new skills.
The algorithm takes the current set of winning states, $Z$.
The postconditions of new skills need to be in $Z$.
As found by the enumeration-based repair, we need two new skills: one with postconditions $\sigma_0 \wedge \sigma_2 \wedge \sigma_5 \wedge \sigma_6 \wedge \sigma_8$ and one with postconditions $\sigma_0 \wedge \sigma_2 \wedge \sigma_3 \wedge \sigma_4 \wedge \sigma_8$.
However, during the synthesis process, all states with either postcondition are removed from $Z$ before the repair process is started.
As a result, we are unable to find a suggestion with the synthesis-based repair process.

When we use the synthesis-based repair as proposed in \citet{pacheck2020finding} without modification, we are able to find suggestions to repair the specification.
However, these suggestions only contain skills which change the truth value of the user define variables $\blueperson$ and $\greenperson$, which is not desired. 
Essentially, they let the robot do its task by enforcing restrictions on how the people are behaving. 
One such suggestion modifies the skills such that when the green plate and blue plate are both clean, if $\blueperson=\true$, it must be $\false$ at the next step if $a_\texttt{green-clean-to-set}$ is not executed to make sure the green plate could be set if needed.
The suggestion also includes extra skills that enforce when the green plate and blue plate are both set and $\blueperson=\false$ and $\greenperson=\true$, $\greenperson$ must become $\false$ so the robot can move the green plate, enabling it to then move the blue plate.


\begin{figure*}
    \centering
    \includegraphics[width=\textwidth]{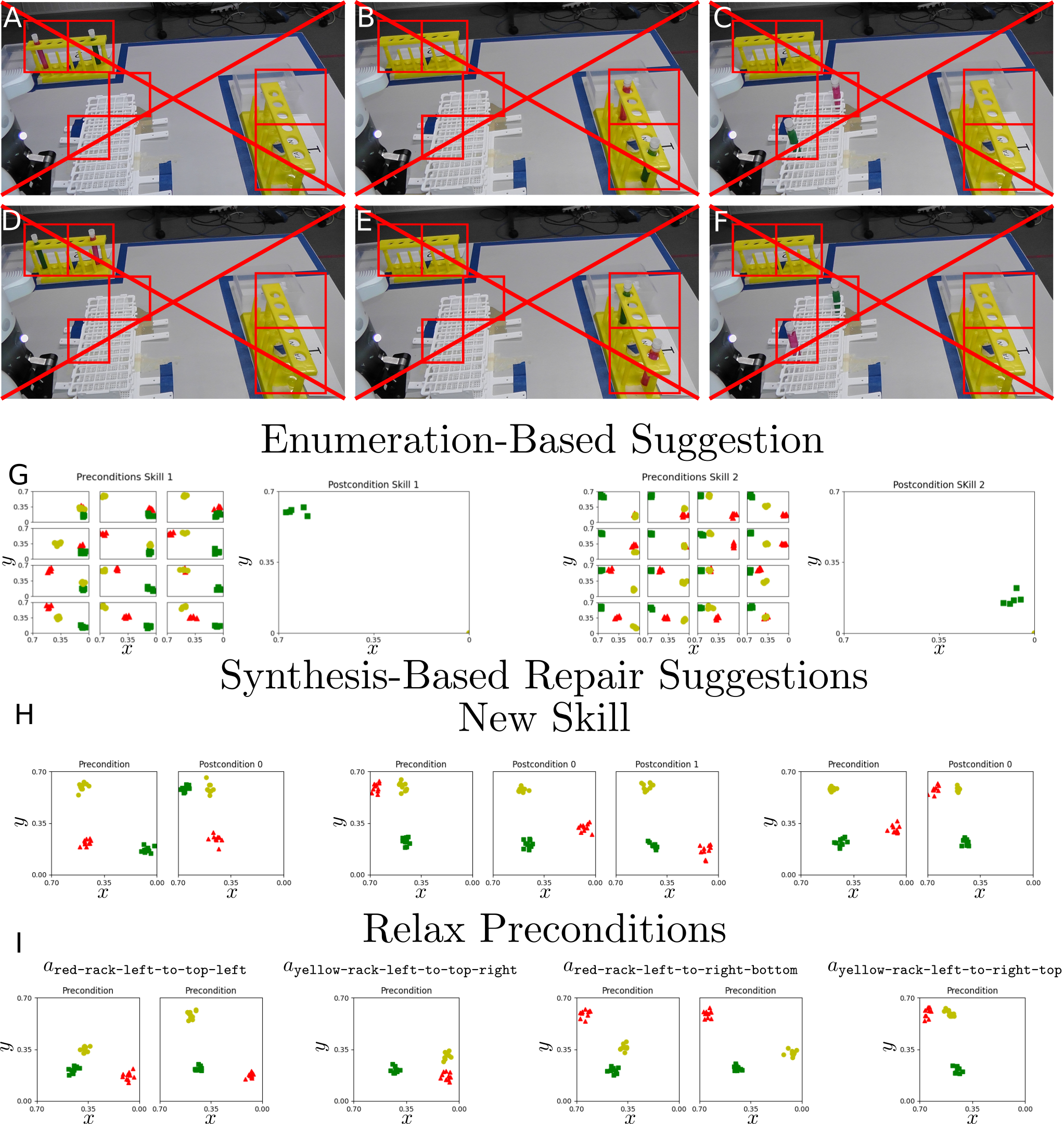}
    \caption{(A-F): Images showing the configurations of the red and green vials that should not occur based on the task specification. In the suggestion plots, the red vials are denoted by red triangles, the green vials by green squares, and the yellow vials by yellow circles. (G): One suggestion given by enumeration-based repair. There are two skills suggested. Note that some of the vials in the preconditions overlap because the preconditions are taken from existing skills. When determining which symbols are contained in the precondition classifier, we do not enforce mutual exclusion of the symbols in the same physical space. Even though these preconditions would violate the mutual exclusion of symbols added in $\phitask$, these states are never visited, so the specification is not violated. (H) Suggestion from the synthesis-based repair. There are three skills suggested. (I) A suggestion from the synthesis-based repair process to relax the preconditions. The repair process suggests relaxing the preconditions of four skills.}
    \label{fig:vaccine_fault_model}
\end{figure*}

\textbf{Kinova Vials:} For the Kinova Vials example, we investigate one unrealizable specification.

\textbf{Unrealizable Specification 6}: 
The user-defined task was the same as in Figure~\ref{fig:vials_liveness}.
We added the constraint that the red and green vials should never be in the white rack at the same time, in addition to the constraint that the red and green vials should never be in yellow racks at the same time.
This additional constraint in $\syssafetytask$ is encoded in Equation~\eqref{eq:vaccine_constraint} and the physical interpretation is shown in Figure~\ref{fig:vaccine_fault_model}(C,F).
\begin{equation}\label{eq:vaccine_constraint}
\begin{split}
    &\square \lnot (\sigma_4 \wedge \sigma_{11}) \wedge \square \lnot (\sigma_5 \wedge \sigma_{10}) \\
    &\square \lnot \bigcirc (\sigma_4 \wedge \sigma_{11}) \wedge \square \lnot \bigcirc (\sigma_5 \wedge \sigma_{10})
\end{split}
\end{equation}

We also add the constraint to $\syssafetytask$ that two vials cannot be in the same physical location at the same time.
We show portion of this constraint in Equation~\eqref{eq:vaccine_only_one_vial_per_loc} and the full constraint in Appendix~\ref{sec:appendix_kinova_vials}.
\begin{equation}\label{eq:vaccine_only_one_vial_per_loc}
    \square \lnot (\sigma_{0} \wedge \sigma_{6}) \wedge \square \lnot (\sigma_{0} \wedge \sigma_{12}) \wedge \square \lnot(\sigma_{1} \wedge \sigma_{7}) \wedge \ldots
\end{equation}
\begin{figure*}
\begin{equation}\label{eq:vaccine_only_one_moves}
\begin{split}
&((\sigma_{6} \leftrightarrow \lnot \bigcirc \sigma_{6}) \vee (\sigma_{7} \leftrightarrow  \lnot \bigcirc \sigma_{7}) \vee (\sigma_{8} \leftrightarrow  \lnot \bigcirc \sigma_{8}) \vee (\sigma_{9} \leftrightarrow  \lnot \bigcirc \sigma_{9}) \vee (\sigma_{10} \leftrightarrow  \lnot \bigcirc \sigma_{10}) \vee (\sigma_{11} \leftrightarrow  \lnot \bigcirc \sigma_{11})) \rightarrow \\ 
& ((\sigma_{0} \leftrightarrow \bigcirc \sigma_{0}) \wedge (\sigma_{1} \leftrightarrow \bigcirc \sigma_{1}) \wedge (\sigma_{2} \leftrightarrow \bigcirc \sigma_{2}) \wedge (\sigma_{3} \leftrightarrow \bigcirc \sigma_{3}) \wedge (\sigma_{4} \leftrightarrow \bigcirc \sigma_{4}) \wedge (\sigma_{5} \leftrightarrow \bigcirc \sigma_{5})) \wedge \\
& ((\sigma_{12} \leftrightarrow \bigcirc \sigma_{12}) \wedge (\sigma_{13} \leftrightarrow \bigcirc \sigma_{13}) \wedge (\sigma_{14} \leftrightarrow \bigcirc \sigma_{14}) \wedge (\sigma_{15} \leftrightarrow \bigcirc \sigma_{15}) \wedge (\sigma_{16} \leftrightarrow \bigcirc \sigma_{16}) \wedge (\sigma_{17} \leftrightarrow \bigcirc \sigma_{17}))
\end{split}
\end{equation}
\end{figure*}
Additionally, we add that only one vial can be moved at a time.
We show the constraint that if the red vial moves, the green and yellow vials cannot move in Equation~\eqref{eq:vaccine_only_one_moves}.
The constraints for the green and yellow vials are similar and we show them in Appendix~\ref{sec:appendix_kinova_vials}.
Without these additional constraints, the suggestions returned are difficult to interpret and unusable because the suggestions involve moving multiple vials at once, which a single arm cannot do.

\textbf{Unrealizable Specification 6 Repair}:
For the enumeration-based repair, we found that $\skills_\textrm{n.o.t} \neq \varnothing$.
However, at least two skills are required to repair the specification.
This necessitates looping through combinations of the skills in $\skills_\textrm{n.o.t}$ and the 48 other partitioned skills.
We stopped the repair process after 7 hours and found 22 suggestions.
One of the suggestions was a skill to move the green vial from the right-bottom location to the top-left location and another skill to move the green vial from the top-left location to the right-bottom location.
Figure~\ref{fig:vaccine_fault_model}G shows this suggestion.

The synthesis-based repair process returns 25 suggestions in 381 seconds.
This is orders of magnitude faster than the enumeration-based repair process.
One of the suggestions is shown in Figure~\ref{fig:vaccine_fault_model}H.
The suggestion proposes three new skills: one with preconditions $\sigma_0 \wedge \sigma_{10} \wedge \sigma_{15}$ and postcondition $\sigma_{2} \wedge \sigma_{10} \wedge \sigma_{15}$, one with preconditions $\sigma_4 \wedge \sigma_8 \wedge \sigma_{15}$ and postconditions $(\sigma_4 \wedge \sigma_6 \wedge \sigma_{15}) \vee (\sigma_4 \wedge \sigma_7 \wedge \sigma_{15})$, and one with preconditions $\sigma_4 \wedge \sigma_7 \wedge \sigma_{15}$ and postcondition $\sigma_4 \wedge \sigma_8 \wedge \sigma_{15}$.
Note that the synthesis-based repair suggestion contains three skills, even though the specification could be repaired with two, while the enumeration-based repair only contains two skills.
This is because the synthesis-based repair has no restrictions on how many skills it can propose and does not attempt to minimize the number of skills suggested.
The enumeration-based repair would consider two skills to repair a specification before considering three, as it would add substantial computation expense.

The synthesis-based repair process also produces suggestions that relax the preconditions of skills.
We show one such suggestion in Figure~\ref{fig:vaccine_fault_model}I.
The suggestion suggests relaxing the preconditions of $a_\texttt{red-rack-left-to-top-left}$ to include states when the red vial is in the right-bottom location, the green vial is in the rack-left, and the yellow vial is in the rack-right or top-right locations.
It also suggests allowing $a_\texttt{yellow-rack-left-to-top-right}$ to be allowed when the yellow vial is in the right-top location, the red vial is in the right-bottom location, and the green vial is in the rack-left location.
The third precondition it suggests relaxing is that for $a_\texttt{red-rack-left-to-right-bottom}$ to include the state when the red vial is in the top-left location, the yellow vial is in the rack-right or right-top locations, and the green vial is in the rack-left location.
It also suggests allowing $a_\texttt{yellow-rack-left-to-right-top}$ to be executed when the yellow vial is in the top-left location, the red vial is in the top-left location, and the green vial is in the rack-left location.
Note that skills involving moving the yellow vial are not strictly necessary to repair the specification, but are included due to the choices made by the synthesis-based repair process when extracting a strategy.

\section{Conclusion}

In this work, we present a framework for automatically encoding the skills of a robot in an \gls{ltl} formula from sensor data.
We provide a task to the robot and generate a strategy to accomplish the task if possible.
If the task is not possible, we show two methods to repair the specifications by providing skill suggestions that would make the task possible.
We demonstrate the process of encoding the skills in an \gls{ltl} formula, an enumeration-based repair process, and a synthesis-based repair process on three examples.

In the future, we plan to extend this work to automatically implement controllers based on the symbolic suggestions.
Additionally, we plan to extend the synthesis-based repair process to handle a larger class of specifications, such as those in the Baxter Plates example or specifications with constraints on what happens during the execution of the skills. 

\section*{Acknowledgments}
This work is supported by the ONR PERISCOPE MURI award N00014-17-1-2699.


\appendix
\section{Appendix}\label{sec:appendix}

We provide the full $\phifixed$ for the Baxter Blocks, Baxter Plates, and Kinova Vials examples in the structuredslugs format \cite{ehlers2016slugs}.
We form the parts of $\phifull$ for each specification by conjuncting each line of the structuredslugs file.
[ENV\_INIT] forms $\envinit$ and [SYS\_INIT] forms $\sysinit$.
We place the temporal operator always ($\square$) in front of each line in [ENV\_TRANS], [SYS\_TRANS], and [SYS\_TRANS\_HARD] before conjuncting them to form $\envsafety$, $\syssafety$, and $\syshard$, respectively.
We prepend each line in [SYS\_LIVENESS] and [ENV\_LIVENESS] with $\square \lozenge$ before conjuncting them to form $\syslive$ and $\envlive$, respectively.
The section [SYS\_TRANS\_HARD] does not appear in a standard structuredslugs file and is included for use with the synthesis-based repair process (Section~\ref{sec:synthesis_base_repair}).
When we evaluate the specifications using Slugs \cite{ehlers2016slugs}, we combine the [SYS\_TRANS\_HARD] section with [SYS\_TRANS].

Indented lines are continuations of a single line and a consequence of the limited column width.

Comments (lines beginning with ''\#'') in the structured slugs file show where the automatically generated portion of the specification from Section~\ref{sec:constant_spec} go.
For each specification, we show the line(s) added to make the specification unrealizable.

\subsection{Baxter Blocks} \label{sec:appendix_baxter_blocks}

\begin{verbatim}
[INPUT]
# 19 automatically generated symbols s0-s18
Switch

[OUTPUT]
# 18 partitioned skills
extra1

[ENV_INIT]
!s0
!s1
!s14
!s15
!s2
!s3
!s4
!s5
!s6
!s8
!Switch
s10
s11
s12
s13
s16
s17
s18
s7
s9

[SYS_INIT]
!a_to_d_4
!a_to_d_5
!a_to_d_6
!a_to_d_7
!a_to_e_11
!a_to_e_12
!a_to_e_13
!a_to_e_14
!b_to_f_19
!c_to_b_10
!d_to_a_15
!d_to_a_16
!d_to_at_8
!d_to_at_9
!e_to_a_17
!e_to_a_18
!e_to_at_2
!e_to_at_3
!f_to_c_0
!g_to_c_1
!extra1

[ENV_TRANS]
# Automatically generated postconditions 
# and mutual exclusion of symbols

[SYS_TRANS]
# Automatically generated preconditions

[SYS_TRANS_HARD]
# Mutual exclusion of skills

# Only one block can move at a time
((s6 <-> !s6') | (s9 <-> !s9') | 
    (s14 <-> !s14') | (s7 <-> !s7') | 
    (s10 <-> !s10') | (s8 <-> !s8')) -> 
    ((s11 <-> s11') & (s3 <-> s3') & 
    (s13 <-> s13') & (s4 <-> s4') & 
    (s12 <-> s12') & (s5 <-> s5') & 
    (s0 <-> s0') & (s17 <-> s17') & 
    (s1 <-> s1') & (s15 <-> s15') & 
    (s18 <-> s18') & (s16 <-> s16') & 
    (s2 <-> s2'))
((s11 <-> !s11') | (s3 <-> !s3') | 
    (s13 <-> !s13') | (s4 <-> !s4') | 
    (s12 <-> !s12') | (s5 <-> !s5')) -> 
    ((s6 <-> s6') & (s9 <-> s9') & 
    (s14 <-> s14') & (s7 <-> s7') & 
    (s10 <-> s10') & (s8 <-> s8') & 
    (s0 <-> s0') & (s17 <-> s17') & 
    (s1 <-> s1') & (s15 <-> s15') & 
    (s18 <-> s18') & (s16 <-> s16') & 
    (s2 <-> s2'))
((s0 <-> !s0') | (s17 <-> !s17') | 
    (s1 <-> !s1') | (s15 <-> !s15') | 
    (s18 <-> !s18') | (s16 <-> !s16') | 
    (s2 <-> !s2')) -> ((s6 <-> s6') & 
    (s9 <-> s9') & (s14 <-> s14') & 
    (s7 <-> s7') & (s10 <-> s10') & 
    (s8 <-> s8') & (s11 <-> s11') & 
    (s3 <-> s3') & (s13 <-> s13') & 
    (s4 <-> s4') & (s12 <-> s12') & 
    (s5 <-> s5'))

# Unrealizable Specification 1
!(s0' & s15')
!(s0 & s15)

# Unrealizable Specification 2
!(c_to_b_10')
!(c_to_b_10)

[SYS_LIVENESS]
Switch -> (s6 & s7 & s10 & s3 & s4 & 
    s5 & s17 & s18 & s16)
!Switch -> (s6 & s7 & s8 & s3 & s4 & 
    s12 & s0 & s1 & s2)

[ENV_LIVENESS]
(g_to_c_1) -> (s0 & s1 & s2)
\end{verbatim}

\subsection{Baxter Plates}\label{sec:appendix_baxter_plates}

\begin{verbatim}
[INPUT]
# 9 automatically generated symbols
# symbol_0 to symbol_8
blue_person
green_person

[OUTPUT]
# 7 automatically partitioned skills
extra1
extra2
extra3

[ENV_INIT]

!symbol_7
!symbol_5
!symbol_4
!symbol_8
symbol_0
symbol_1
symbol_2
symbol_3
symbol_6

[SYS_INIT]
!blue_dirty_to_clean_partition_0_3
!green_clean_to_set_partition_0_1
!green_set_to_dirty_partition_0_2
!extra1
!extra2
!blue_set_to_dirty_partition_0_5
!extra3
!green_clean_to_set_partition_0_0
!blue_clean_to_set_partition_0_4
!green_dirty_to_clean_partition_0_6

[ENV_TRANS]
# Automatically generated postconditions 
# and mutual exclusion of symbols

[SYS_TRANS]
# Automatically generated preconditions

[SYS_TRANS_HARD]
# Mutual exclusion of skills

# Unrealizable Specification 5
blue_person & blue_person' & 
    ((symbol_0 & symbol_1 & symbol_2 & 
    symbol_5 & symbol_6) | 
    (symbol_0 & symbol_1 & symbol_5 & 
    symbol_6 & symbol_7) | 
    (symbol_0 & symbol_2 & symbol_5 & 
    symbol_6 & symbol_8)) -> 
    ((symbol_0' & symbol_1' & symbol_2' & 
    symbol_5' & symbol_6') | 
    (symbol_0' & symbol_1' & symbol_5' & 
    symbol_6' & symbol_7') | 
    (symbol_0' & symbol_2' & symbol_5' & 
    symbol_6' & symbol_8'))
green_person & green_person' & 
    ((symbol_0 & symbol_2 & symbol_3 & 
    symbol_6 & symbol_8) | 
    (symbol_0 & symbol_2 & symbol_3 & 
    symbol_4 & symbol_8) | 
    (symbol_0 & symbol_2 & symbol_5 & 
    symbol_6 & symbol_8)) -> 
    ((symbol_0' & symbol_2' & symbol_3' & 
    symbol_6' & symbol_8') | 
    (symbol_0' & symbol_2' & symbol_3' & 
    symbol_4' & symbol_8') | 
    (symbol_0' & symbol_2' & symbol_5' & 
    symbol_6' & symbol_8'))

[SYS_LIVENESS]
blue_person -> 
    ((symbol_0 & symbol_1 & symbol_2 & 
    symbol_5 & symbol_6) | 
    (symbol_0 & symbol_1 & symbol_5 & 
    symbol_6 & symbol_7) | 
    (symbol_0 & symbol_2 & symbol_5 & 
    symbol_6 & symbol_8))
green_person ->
    ((symbol_0 & symbol_2 & symbol_3 & 
    symbol_6 & symbol_8) | 
    (symbol_0 & symbol_2 & symbol_3 & 
    symbol_4 & symbol_8) | 
    (symbol_0 & symbol_2 & symbol_5 & 
    symbol_6 & symbol_8))
!blue_person -> 
    !((symbol_0 & symbol_1 & symbol_2 & 
    symbol_5 & symbol_6) | 
    (symbol_0 & symbol_1 & symbol_5 & 
    symbol_6 & symbol_7) | 
    (symbol_0 & symbol_2 & symbol_5 & 
    symbol_6 & symbol_8))
!green_person -> 
    !((symbol_0 & symbol_2 & symbol_3 & 
    symbol_6 & symbol_8) | 
    (symbol_0 & symbol_2 & symbol_3 & 
    symbol_4 & symbol_8) | 
    (symbol_0 & symbol_2 & symbol_5 & 
    symbol_6 & symbol_8))

[ENV_LIVENESS]
green_clean_to_set_partition_0_1 -> 
    (symbol_8)
blue_clean_to_set_partition_0_4 -> 
    (symbol_5)
\end{verbatim}

\subsection{Kinova Vials}\label{sec:appendix_kinova_vials}

\begin{verbatim}
[INPUT]
# 18 automatically generated symbols
# s0 to s17
react

[OUTPUT]
# 48 skills
extra1
extra2
extra3

[ENV_INIT]

!s1
!s10
!s11
!s12
!s13
!s14
!s16
!s17
!s2
!s3
!s4
!s5
!s6
!s7
!s9
!react
s0
s15
s8

[SYS_INIT]

!green_rack_left_to_right_bottom_0
!green_rack_left_to_right_top_1
!green_rack_left_to_top_left_2
!green_rack_left_to_top_right_3
!green_rack_right_to_right_bottom_4
!green_rack_right_to_right_top_5
!green_rack_right_to_top_left_6
!green_rack_right_to_top_right_7
!green_right_bottom_to_rack_left_8
!green_right_bottom_to_rack_right_9
!green_right_top_to_rack_left_10
!green_right_top_to_rack_right_11
!green_top_left_to_rack_left_12
!green_top_left_to_rack_right_13
!green_top_right_to_rack_left_14
!green_top_right_to_rack_right_15
!red_rack_left_to_right_bottom_16
!red_rack_left_to_right_top_17
!red_rack_left_to_top_left_18
!red_rack_left_to_top_right_19
!red_rack_right_to_right_bottom_20
!red_rack_right_to_right_top_21
!red_rack_right_to_top_left_22
!red_rack_right_to_top_right_23
!red_right_bottom_to_rack_left_24
!red_right_bottom_to_rack_right_25
!red_right_top_to_rack_left_26
!red_right_top_to_rack_right_27
!red_top_left_to_rack_left_28
!red_top_left_to_rack_right_29
!red_top_right_to_rack_left_30
!red_top_right_to_rack_right_31
!yellow_rack_left_to_right_bottom_32
!yellow_rack_left_to_right_top_33
!yellow_rack_left_to_top_left_34
!yellow_rack_left_to_top_right_35
!yellow_rack_right_to_right_bottom_36
!yellow_rack_right_to_right_top_37
!yellow_rack_right_to_top_left_38
!yellow_rack_right_to_top_right_39
!yellow_right_bottom_to_rack_left_40
!yellow_right_bottom_to_rack_right_41
!yellow_right_top_to_rack_left_42
!yellow_right_top_to_rack_right_43
!yellow_top_left_to_rack_left_44
!yellow_top_left_to_rack_right_45
!yellow_top_right_to_rack_left_46
!yellow_top_right_to_rack_right_47
!extra1
!extra2
!extra3

[ENV_TRANS]
# Automatically generated postconditions 
# and mutual exclusion of symbols

[SYS_TRANS]
# Automatically generated preconditions

[SYS_TRANS_HARD]
# Mutual exclusion of skills

# Realizable Specification
# The red and green vials should not be 
# in the same yellow rack
!(s0' & s7')
!(s1' & s6')
!(s2' & s9')
!(s3' & s8')
!(s0 & s7)
!(s1 & s6)
!(s2 & s9)
!(s3 & s8)
# Only one vial can be in a location
!(s0' & s6')
!(s0' & s12')
!(s6' & s12')
!(s1' & s7')
!(s1' & s13')
!(s7' & s13')
!(s2' & s8')
!(s2' & s14')
!(s8' & s14')
!(s3' & s9')
!(s3' & s15')
!(s9' & s15')
!(s4' & s10')
!(s4' & s16')
!(s10' & s16')
!(s5' & s11')
!(s5' & s17')
!(s11' & s17')
!(s0 & s6)
!(s0 & s12)
!(s6 & s12)
!(s1 & s7)
!(s1 & s13)
!(s7 & s13)
!(s2 & s8)
!(s2 & s14)
!(s8 & s14)
!(s3 & s9)
!(s3 & s15)
!(s9 & s15)
!(s4 & s10)
!(s4 & s16)
!(s10 & s16)
!(s5 & s11)
!(s5 & s17)
!(s11 & s17)
# Only one vial can move at a time
((s6 <-> !s6') | (s8 <-> !s8') | 
    (s7 <-> !s7') | (s9 <-> !s9') | 
    (s11 <-> !s11') | (s10 <-> !s10')) -> 
    ((s0 <-> s0') & (s1 <-> s1') & 
    (s2 <-> s2') & (s3 <-> s3') & 
    (s4 <-> s4') & (s5 <-> s5')) & 
    ((s12 <-> s12') & (s13 <-> s13') & 
    (s14 <-> s14') & (s15 <-> s15') & 
    (s16 <-> s16') & (s17 <-> s17'))
((s0 <-> !s0') | (s1 <-> !s1') | 
    (s2 <-> !s2') | (s3 <-> !s3') | 
    (s4 <-> !s4') | (s5 <-> !s5')) -> 
    ((s6 <-> s6') & (s8 <-> s8') & 
    (s7 <-> s7') & (s9 <-> s9') & 
    (s11 <-> s11') & (s10 <-> s10')) & 
    ((s12 <-> s12') & (s13 <-> s13') & 
    (s14 <-> s14') & (s15 <-> s15') & 
    (s16 <-> s16') & (s17 <-> s17'))
((s12 <-> !s12') | (s13 <-> !s13') | 
    (s14 <-> !s14') | (s15 <-> !s15') | 
    (s16 <-> !s16') | (s17 <-> !s17')) 
    -> ((s6 <-> s6') & (s8 <-> s8') & 
    (s7 <-> s7') & (s9 <-> s9') & 
    (s11 <-> s11') & (s10 <-> s10')) & 
    ((s0 <-> s0') & (s1 <-> s1') & 
    (s2 <-> s2') & (s3 <-> s3') & 
    (s4 <-> s4') & (s5 <-> s5'))

# Unrealizable Specification 6
!(s4' & s11')
!(s5' & s10')
!(s4 & s11)
!(s5 & s10)
 
[SYS_LIVENESS]
react -> (s8 & s15 & s0)
!react -> (s2 & s13 & s6)

[ENV_LIVENESS]
\end{verbatim}

\end{document}